\pdfoutput=1

\documentclass[11pt]{article}
\usepackage{colortbl}
\usepackage[table]{xcolor}
\usepackage[final]{acl}

\usepackage{times}
\usepackage{latexsym}

\usepackage[T1]{fontenc}

\usepackage[utf8]{inputenc}

\usepackage{microtype}

\usepackage{inconsolata}

\usepackage{setspace}

\usepackage{algorithm}
\usepackage{algorithmic}

\usepackage{graphicx}
\usepackage{times}
\usepackage{latexsym}
\usepackage[T1]{fontenc}
\usepackage[utf8]{inputenc}
\usepackage{pifont}

\usepackage{soul}
\usepackage{microtype}
\usepackage{tabularx}
\usepackage{xspace}
\usepackage{makecell}
\usepackage{tipa}
\usepackage{graphicx}
\usepackage{booktabs}
\usepackage{multirow}
\usepackage{colortbl}
\usepackage{xcolor}
\usepackage{caption}
\usepackage{amsmath} 
\usepackage{arydshln}
\usepackage{amssymb}  
\usepackage{tabularx}
\usepackage{color}
\usepackage{bbm}
\usepackage{svg}
\usepackage{marvosym}
\usepackage[dvipsnames]{xcolor}
\usepackage{tcolorbox}
\tcbuselibrary{skins,breakable}

\usepackage{subcaption}
\newcommand{\modelLogo}[1]{\raisebox{-0.12em}{\includegraphics[height=0.95em]{figs/logo/#1.pdf}}\hspace{0.25em}}
\newcommand{\chatgptLogo}{\modelLogo{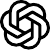}}
\newcommand{\llamaLogo}{\modelLogo{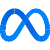}}
\newcommand{\qwenLogo}{\modelLogo{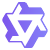}}
\newcommand{\graniteLogo}{\modelLogo{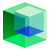}}
\newcommand{\geminiLogo}{\modelLogo{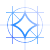}}
\newcommand{\best}[1]{\textbf{#1}}
\newcommand{\secondbest}[1]{\underline{#1}}
\newcommand{\flatAvg}[1]{\textcolor{blue}{\textbf{#1}}}
\newcommand{\nestedAvg}[1]{\textcolor{blue}{\textbf{#1}}}
\definecolor{myrowblue}{HTML}{EAF4FF}
\definecolor{outerboxframe}{RGB}{180, 180, 180}

\title{Assessment of Generative Named Entity Recognition in the Era of \\Large Language Models}

\author{
    Qi Zhan, 
    Yile Wang\thanks{Corresponding author.},
    Hui Huang \\
    Guangdong Provincial Key Laboratory of Visual Media and Multidimensional Intelligence\\CSSE, Shenzhen University \\
    \texttt{qzhan65@gmail.com, wangyile@szu.edu.cn, hhzhiyan@gmail.com}
}

\begin{document}
\maketitle

\begin{abstract}
Named entity recognition (NER) is evolving from a sequence labeling task into a generative paradigm with the rise of large language models (LLMs). We conduct a systematic evaluation of open-source LLMs on both flat and nested NER tasks. We investigate several research questions including the performance gap between generative NER and traditional NER models, the impact of output formats, whether LLMs rely on memorization, and the preservation of general capabilities after fine-tuning. Through experiments across eight LLMs of varying scales and four standard NER datasets, we find that: (1) With parameter-efficient fine-tuning and structured formats like inline bracketed or XML, open-source LLMs achieve performance competitive with traditional encoder-based models and surpass decoder-based LLMs with in-context learning techniques; (2) The NER capability of LLMs stems from instruction-following and generative power, not mere memorization of entity-label pairs; and (3) Applying NER instruction tuning has minimal impact on general capabilities of LLMs, even improving performance on datasets like DROP by 25.50 to 45.32 F1 points due to enhanced entity understanding. These findings demonstrate that generative NER with LLMs is a promising, user-friendly alternative to traditional methods. We release the data and code at \href{https://github.com/szu-tera/LLMs4NER}{https://github.com/szu-tera/LLMs4NER}.
\end{abstract}

\section{Introduction}
Named entity recognition (NER) is a fundamental task in natural language processing, widely used in downstream applications such as relation extraction, knowledge graph construction, information retrieval, and question answering systems. As a typical sequence labeling task, approaches based on traditional auto-encoding pre-trained models have achieved strong results~\cite{devlin-etal-2019-bert,li-etal-2020-unified,lu-etal-2022-unified,yan-etal-2023-embarrassingly}.

\begin{figure}[t]
    \centering
    \includegraphics[width=1.0\columnwidth]{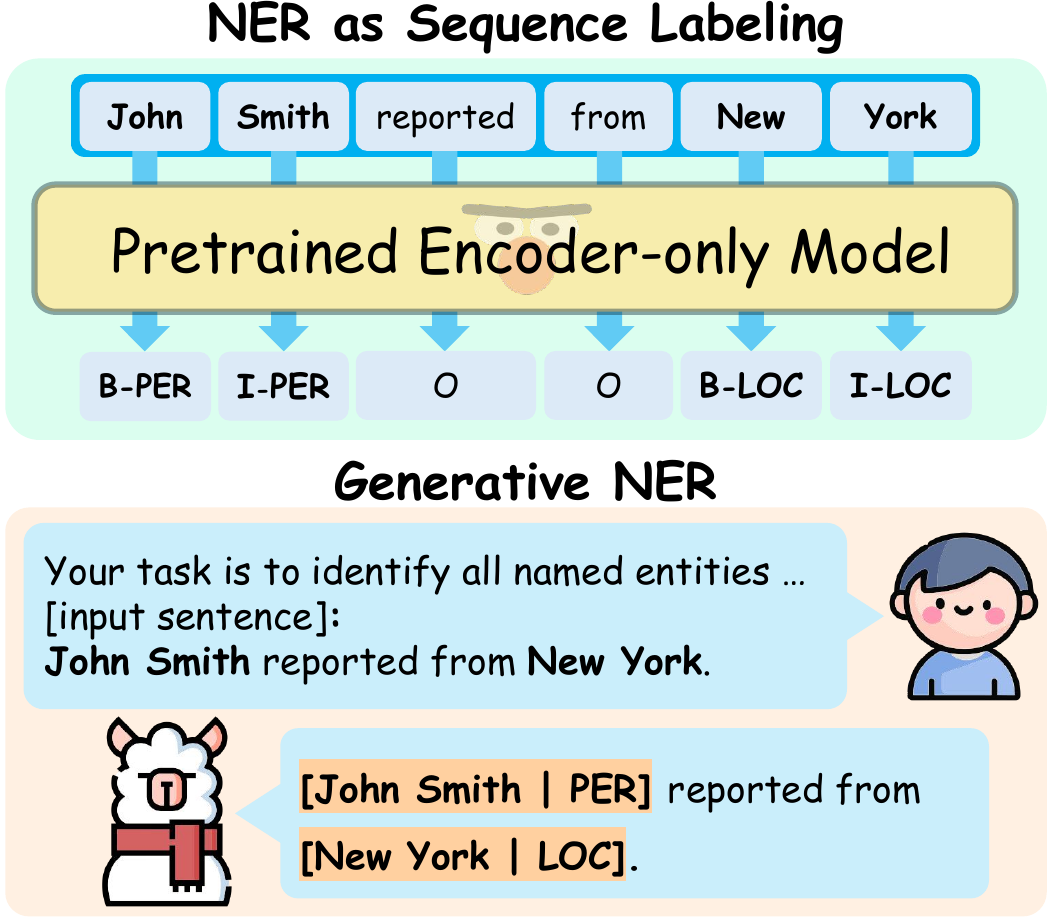} 
    \caption{Comparison between traditional NER based on pre-trained models (top) and generative NER based on large language models (bottom).}
    \label{fig:framework_comparison}
\end{figure}

The development of generative pre-trained models~\cite{radford2019language,lewis-etal-2020-bart} and large language models (LLMs; \citealp{NEURIPS2020_1457c0d6,openaichatgpt,achiam2023gpt,liu2024deepseek}) is gradually shifting the traditional NER task toward a unified generative paradigm. As shown in Figure~\ref{fig:framework_comparison}, LLMs sequentially generate user-provided sentences along with their corresponding entity labels in textual form. This approach demonstrates strong interactivity, generalization, and user-friendliness. Recent studies have conducted preliminary investigations into NER based on LLMs by crafting specific prompts~\cite{xie-etal-2023-empirical,wei2023zero,kim-etal-2024-exploring,wang-etal-2025-gpt}, also achieving promising results.

In this work, we aim to further explore the following research questions: \textbf{In terms of performance, is generative NER with open-source LLMs a reliable approach compared with advanced pre-trained NER models?} Most of the aforementioned studies test the performance of closed-source models like ChatGPT~\cite{openaichatgpt} for zero-shot NER, and there is currently a lack of systematic investigation into the effectiveness of fine-tuned open-source LLMs, such as advanced LLaMA~\cite{grattafiori2024llama}, Qwen~\cite{yang2025qwen3}, Granite~\cite{granite2025} and Gemma~\cite{gemmateam2025gemma3technicalreport} models. Specifically, we seek to examine whether generative LLMs represent a reliable approach and whether there is a performance gap compared to traditional pre-trained model methods on both flat and nested NER tasks.

\textbf{How do different output formats affect the performance of LLMs in generative NER?} The performance of LLMs is highly sensitive to the specific instructions given by users. While recent studies have touched upon format designs~\cite{kim-etal-2024-exploring,lv2025unified}, we systematically explore a more diverse set of output templates as a supplement to the aforementioned work and investigate their specific impact on NER performance. Specifically, we design five distinct output formats (Inline Bracketed, Inline XML, Category-grouped JSON, Occurrence-based JSON, and Offset-based JSON) to compare their impact and investigate whether the performance of different models remains consistent across these various formats.

\textbf{Do LLMs rely on memorization of entity labels to solve NER tasks?} Named entities are prevalent in corpora and are incorporated into the training process of LLMs. Moreover, studies indicate that portions of certain datasets have been leaked during LLM training, casting doubt on the reliability of evaluation results~\cite{sainz-etal-2023-nlp,dong-etal-2024-generalization,balloccu-etal-2024-leak}. Therefore, we aim to investigate whether the generative NER capability of LLMs stems merely from memorizing entity-label mappings, rather than from genuinely learning the features of entities and accurately identifying and labeling them. Specifically, we evaluate LLMs under few-shot prompting without fine-tuning and replace standard text labels with arbitrary symbols to test whether models rely on memorized entity-label correlations.

\textbf{Do LLMs preserve the original capabilities after instruction tuning for NER?} One advantage of LLMs is their strong generalization capability, which enables them to perform well across a wide range of tasks. We investigate whether the original abilities of LLMs are affected after instruction tuning on NER tasks. To this end, we compare the performance of the LLM before and after fine-tuning on multiple datasets, including HellaSwag~\cite{zellers-etal-2019-hellaswag}, DROP~\cite{dua-etal-2019-drop}, GSM8K~\cite{cobbe2021trainingverifierssolvemath}, MMLU~\cite{hendrycks2021measuring}, and TruthfulQA~\cite{lin-etal-2022-truthfulqa}.

In this study, we conduct experiments using eight LLMs with varying sizes and versions of LLaMA, Qwen, Granite, and Gemma across four standard datasets, including two flat NER datasets CoNLL2003~\cite{tjong-kim-sang-de-meulder-2003-introduction} and OntoNotes5.0~\cite{pradhan-etal-2013-towards}, and two nested NER datasets ACE2005~\cite{walker2006ace} and GENIA~\cite{GENIA}. Empirically, the answers to the above questions are as follows:

1) With parameter-efficient fine-tuning and structured output formats like inline bracketed or inline XML, LLMs match the performance of traditional encoder-based NER models and exceed that of closed-source LLMs like GPT-3. They excel particularly in flat NER within general domains, though further progress is needed for nested NER in domains like biomedicine.

2) The strong performance of LLMs in generative NER tasks stems primarily from their powerful instruction-following and text generation capabilities, rather than from merely memorizing entity-label correlations from training corpora. Error analysis indicates that LLMs tend to generate wrong types, which stems from a misalignment between the rich intrinsic knowledge and the detailed annotation guidelines of datasets.

3) LLMs can retain most of their original capabilities under parameter-efficient fine-tuning for NER tasks, with performance on tasks staying stable or slightly decreasing while improving significantly on benchmarks like DROP~\cite{dua-etal-2019-drop} by 25.50 to 45.32 F1 points. Considering their user-friendly natural language interface, which lowers the barrier for non-experts, generative NER systems hold the potential to gradually supplant traditional sequence labeling based NER models.

\begin{table*}[t!]
    \centering
    \resizebox{2.08\columnwidth}{!}{
	\begin{tabular}{ll}
    \toprule
    \textbf{Types}&\textbf{Examples of Labeled Output}\\
    \midrule
        \multirow{2}*{Inline Bracketed}
        &This is the first demonstration of a specific interaction with [PU.1 \textcolor{blue}{| \textsc{Protein}}]\\
        & on a [myeloid [PU.1 \textcolor{blue}{| \textsc{Protein}}] binding site \textcolor{magenta}{| \textsc{DNA}}].\\
    
    \midrule

    \multirow{2}*{Inline XML}
    & This is the first demonstration of a specific interaction with \textcolor{blue}{<\textsc{Protein}>}PU.1\textcolor{blue}{</\textsc{Protein}>} \\
    & on a \textcolor{magenta}{<\textsc{DNA}>}myeloid \textcolor{blue}{<\textsc{Protein}>}PU.1\textcolor{blue}{</\textsc{Protein}>} binding site\textcolor{magenta}{</\textsc{DNA}>}. \\
    \midrule
    \multirow{2}*{Category-grouped JSON} 
    & \{\textcolor{magenta}{``\textsc{DNA}''}: [``myeloid PU.1 binding site''], \textcolor{blue}{``\textsc{Protein}''}: [``PU.1'', ``PU.1''], \\
    & \phantom{\{} ``\textsc{Cell\_line}'': [], ``\textsc{Cell\_type}'': [], ``\textsc{RNA}'': []\} \\
    \midrule
        \multirow{3}*{Occurrence-based JSON}
    & [\{``text'': ``PU.1'', ``label'': \textcolor{blue}{``\textsc{Protein}''}, ``occurrence\_index'': 1\}, \\
    & \phantom{[}\{``text'': ``myeloid PU.1 binding site'', ``label'': \textcolor{magenta}{``\textsc{DNA}''}, ``occurrence\_index'': 1\}, \\
    & \phantom{[}\{``text'': ``PU.1'', ``label'': \textcolor{blue}{``\textsc{Protein}''}, ``occurrence\_index'': 2\}] \\
    
    \midrule
    \multirow{3}*{Offset-based JSON} 
    & [\{``text'': ``PU.1'', ``label'': \textcolor{blue}{``\textsc{Protein}''}, ``start'': 63, ``end'': 67\},  \\
    & \phantom{[}\{``text'': ``myeloid PU.1 binding site'', ``label'': \textcolor{magenta}{``\textsc{DNA}''}, ``start'': 73, ``end'': 98\}, \\
    & \phantom{[}\{``text'': ``PU.1'', ``label'': \textcolor{blue}{``\textsc{Protein}''}, ``start'': 81, ``end'': 85\}] \\
    \bottomrule
	\end{tabular}}
	\caption{Five types of output formats and examples for both flat and nested named entities.}
	\label{table:format}
\end{table*}

\section{Related Work}
\noindent\textbf{NER with Traditional and Pre-trained Models.} Traditional NER methods utilize CNN or LSTM approaches for NER~\cite{ma-hovy-2016-end,yang-etal-2018-design}, where no prior knowledge is used before model training. Subsequently, pre-trained models represented by BERT~\cite{devlin-etal-2019-bert} further enhanced the performance of NER, gradually becoming solid backbone models and reaching state-of-the-art levels~\cite{li-etal-2020-unified,Li_Fei_Liu_Wu_Zhang_Teng_Ji_Li_2022,lu-etal-2022-unified,yan-etal-2023-embarrassingly}. Overall, these methods treat NER as a token-level sequence labeling task, utilizing auto-encoding models for text modeling. \citet{cui-etal-2021-template} and \citet{ma-etal-2022-template} explore the use of generative BART~\cite{lewis-etal-2020-bart} model for NER, gradually transforming NER from sequence labeling into a sequence-to-sequence generation task, and shifting NER models from auto-encoding representation models to auto-regressive generative models. As a result, the generative NER approach has also demonstrated promising results, while offering good flexibility and generalization. 

\noindent\textbf{Generative NER with Large Language Models.} Recent studies employ advanced LLMs for generative NER. \citet{xie-etal-2023-empirical} and \citet{wei2023zero} preliminarily explore the performance of ChatGPT on zero-shot NER. \citet{kim-etal-2024-exploring} test the performance of ChatGPT and LLaMA on nested NER by designing fixed templates. \citet{zhou2024universalner} distill knowledge from ChatGPT for NER and \citet{wang-etal-2025-gpt} investigate the effectiveness of GPT-3 in NER through in-context learning~\cite{NEURIPS2020_1457c0d6}. These works primarily focus on evaluating zero-shot or few-shot performance of LLMs. However, a comprehensive evaluation and comparison of LLMs for NER tasks remains lacking. We systematically compare the performance of fine-tuned LLMs with traditional state-of-the-art NER models, analyze the impact of different output formats, and explore the memorization and learning capabilities of LLMs for NER tasks, thereby addressing this gap and serving as a complement to the aforementioned research efforts.

\begin{table}[t!]
    \centering
    \resizebox{1\columnwidth}{!}{
	\begin{tabular}{lcccrrr}
    \toprule
    \textbf{Datasets}&\textbf{Types}&\textbf{Domain}&\#\textbf{Labels} &\#\textbf{Train}&\#\textbf{Dev} & \#\textbf{Test}\\
    \midrule
    CoNLL2003 & Flat&General & 4&  14,041&  3,250  &3,453\\
    OntoNotes5.0& Flat &General& 18&  59,924 &  8,528 &8,262\\
    ACE2005&Nested&General& 7& 7,299&  971  &1,060\\
    GENIA& Nested&Biomedical& 5&14,835&  1,854   &1,854\\
    \bottomrule
	\end{tabular}}
	\caption{Statistics of four NER datasets we used.}
	\label{table:statistics}
\end{table}

\noindent\textbf{Recent Exploration on NER.} The advancement of LLMs has given rise to new paradigms for NER. \citet{Huang_Chen_Huang_Lin_Qin_2026} employ reinforcement learning to implement NER based on long chain-of-thought reasoning. \citet{10.1145/3696410.3714923} propose a multi-agent paradigm for zero-shot NER. \citet{Mu_Ning_Zhao_Zhang_2026} also propose an agent-style framework for multi-domain low-resource NER. These approaches offer new directions for future NER systems.

\section{Instructions, Output Format, and Performance of Generative NER}
\subsection{Settings}
\noindent\textbf{Instructions and Output Format.} Prompt design is crucial for LLM fine-tuning. It mainly includes the design of instructions that specify the tasks the model needs to complete, and the output format used for answer generation, extraction, and validation. The complete instructions are provided in Appendix~\ref{app:prompt} for reproducibility.

As listed in Table~\ref{table:format}, we compare five different output formats. \textbf{Inline Bracketed} and \textbf{Inline XML} represent an embedded paradigm where entity boundaries and types are inserted directly into the original sentence using brackets or XML tags, yielding natural-language-like outputs that preserve the original text structure. The JSON-based formats treat NER as a structured information extraction task. \textbf{Category-grouped JSON} organizes entities by their label keys, which can only handle the entity-label relationship while ambiguity exists due to the lack of positional information. \textbf{Occurrence-based JSON} records an occurrence order index for each entity, providing a certain level of positional reference, while \textbf{Offset-based JSON} demands precise character-level start and end positions, whose fine-grained positional alignment poses a greater challenge for generative models.

\noindent\textbf{Datasets.} For in-domain evaluation, to fairly compare with pre-trained model based methods, we employ standard datasets and do not use external instruction data such as in UniversalNER~\cite{zhou2024universalner}. For traditional flat NER, we use CoNLL2003~\cite{tjong-kim-sang-de-meulder-2003-introduction} and OntoNotes5.0~\cite{pradhan-etal-2013-towards} in general domain. For more challenging nested NER, we use ACE2005\footnote{\url{https://catalog.ldc.upenn.edu/LDC2006T06}}~\cite{walker2006ace} in general domain and GENIA~\cite{GENIA} in molecular biology domain. The statistics of datasets are shown in Table~\ref{table:statistics}.

\noindent\textbf{Baselines.} For pre-trained NER models, we compare {BERT-Tagger}~\cite{devlin-etal-2019-bert}, {GNN-SL}~\cite{wang-etal-2023-gnn}, {ACE + Fine-tune}~\cite{wang-etal-2021-automated}, {BERT-MRC+DSC}~\cite{li-etal-2020-dice}, {BERT-MRC}~\cite{li-etal-2020-unified}, {BINDER}~\cite{zhang2023optimizing}, {W2NER}~\cite{Li_Fei_Liu_Wu_Zhang_Teng_Ji_Li_2022}, and {Biaffine+CNN}~\cite{yan-etal-2023-embarrassingly}. See Appendix~\ref{app:baselines} for brief introductions of these methods.

We also compare GPT-NER~\cite{wang-etal-2025-gpt}, which uses closed-source GPT-3 for generative NER. Three settings are used including zero-shot (\textit{ZS}), zero-shot with entity verification (\textit{ZS w/ Ver.}), and few-shot with entity verification (\textit{FS w/ Ver.}).

\noindent\textbf{Models.} We use representative open-source models from four LLM families: LLaMA~\cite{grattafiori2024llama}, Qwen~\cite{yang2025qwen3}, Granite~\cite{granite2025}, and Gemma~\cite{gemmateam2025gemma3technicalreport}. Specifically, we evaluate versions including LLaMA3.1, LLaMA3.2, Qwen2.5, Qwen3, Granite3.3, Granite4, Gemma2, and Gemma3, covering lightweight (3B$\sim$4B) and medium-sized (7B$\sim$9B) parameter scales. 

\begin{table}[t!]
    \scriptsize
    \centering
    \setlength{\tabcolsep}{3pt}
    \renewcommand{\arraystretch}{1.05}
    \begin{tabularx}{1\columnwidth}{@{}>{\raggedright\arraybackslash}p{0.46\columnwidth}>{\centering\arraybackslash}X>{\centering\arraybackslash}X>{\centering\arraybackslash}X@{}}
        \toprule
        \textbf{Models} & \textbf{CoNLL2003} & \textbf{OntoNotes5.0} & \textbf{Average} \\
        \midrule
        \addlinespace[1pt]
        \rowcolor{gray!20}
        \multicolumn{4}{c}{\textit{Pre-Trained NER Models}} \\
        BERT-Tagger~\cite{devlin-etal-2019-bert} & 91.70 & 89.16 & 90.43 \\
        GNN-SL~\cite{wang-etal-2023-gnn} & 93.20 & \secondbest{91.39} & 92.30 \\
        ACE + Fine-tune~\cite{wang-etal-2021-automated} & 93.64 & \secondbest{91.39} & 92.52 \\
        BERT-MRC+DSC~\cite{li-etal-2020-dice} & 93.33 & \best{92.07} & 92.70 \\
        BERT-MRC~\cite{li-etal-2020-unified} & 93.04 & 91.11 & 92.08 \\
        \midrule
        \addlinespace[1pt]
        \rowcolor{gray!20}
        \multicolumn{4}{c}{\textit{GPT-NER}~\cite{wang-etal-2025-gpt}} \\
        \chatgptLogo GPT-3 (\textit{ZS}) & 89.97 & 81.73 & 85.85 \\
        \chatgptLogo GPT-3 (\textit{ZS w/ Ver.}) & 90.62 & 82.08 & 86.35 \\
        \chatgptLogo GPT-3 (\textit{FS w/ Ver.}) & 90.91 & 82.20 & 86.56 \\
        \midrule
        \addlinespace[1pt]
        \rowcolor{gray!20}
        \multicolumn{4}{c}{\textit{Open-Source LLMs (Inline Bracketed)}} \\
        \llamaLogo LLaMA3.1-8B & \best{93.85} & 91.28 & 92.57 \\
        \llamaLogo LLaMA3.2-3B & 93.54 & 89.19 & 91.37 \\
        \qwenLogo Qwen2.5-7B & 93.73 & 90.97 & 92.35 \\
        \qwenLogo Qwen3-4B & 93.37 & 89.30 & 91.34 \\
        \graniteLogo Granite3.3-8B & 93.76 & 90.31 & 92.04 \\
        \graniteLogo Granite4-3B & 93.15 & 89.82 & 91.49 \\
        \geminiLogo Gemma2-9B & 93.64 & 90.91 & 92.28 \\
        \geminiLogo Gemma3-4B & 93.68 & 90.09 & 91.89 \\
        \hdashline
        \rowcolor{myrowblue}
        \textbf{Average} & 93.59 & 90.23 & \flatAvg{91.91} \\
        \midrule
        \addlinespace[1pt]
        \rowcolor{gray!20}
        \multicolumn{4}{c}{\textit{Open-Source LLMs (Inline XML)}} \\
        \llamaLogo LLaMA3.1-8B & \secondbest{93.83} & 90.94 & 92.39 \\
        \llamaLogo LLaMA3.2-3B & 93.06 & 89.30 & 91.18 \\
        \qwenLogo Qwen2.5-7B & 93.59 & 90.79 & 92.19 \\
        \qwenLogo Qwen3-4B & 93.15 & 88.96 & 91.06 \\
        \graniteLogo Granite3.3-8B & 93.33 & 89.95 & 91.64 \\
        \graniteLogo Granite4-3B & 93.07 & 89.46 & 91.27 \\
        \geminiLogo Gemma2-9B & 93.59 & 90.99 & 92.29 \\
        \geminiLogo Gemma3-4B & 93.22 & 90.15 & 91.69 \\
        \hdashline
        \rowcolor{myrowblue}
        \textbf{Average} & 93.36 & 90.07 & \flatAvg{91.71} \\
        \midrule
        \addlinespace[1pt]
        \rowcolor{gray!20}
        \multicolumn{4}{c}{\textit{Open-Source LLMs (Category-grouped JSON)}} \\
        \llamaLogo LLaMA3.1-8B & 92.56 & 89.84 & 91.20 \\
        \llamaLogo LLaMA3.2-3B & 92.41 & 87.34 & 89.88 \\
        \qwenLogo Qwen2.5-7B & 92.74 & 89.48 & 91.11 \\
        \qwenLogo Qwen3-4B & 92.47 & 87.67 & 90.07 \\
        \graniteLogo Granite3.3-8B & 92.97 & 88.98 & 90.98 \\
        \graniteLogo Granite4-3B & 91.78 & 88.44 & 90.11 \\
        \geminiLogo Gemma2-9B & 93.48 & 90.54 & 92.01 \\
        \geminiLogo Gemma3-4B & 92.53 & 89.17 & 90.85 \\
        \hdashline
        \rowcolor{myrowblue}
        \textbf{Average} & 92.62 & 88.93 & \flatAvg{90.78} \\
        \midrule
        \addlinespace[1pt]
        \rowcolor{gray!20}
        \multicolumn{4}{c}{\textit{Open-Source LLMs (Occurrence-based JSON)}} \\
        \llamaLogo LLaMA3.1-8B & 93.54 & 90.52 & 92.03 \\
        \llamaLogo LLaMA3.2-3B & 93.58 & 87.64 & 90.61 \\
        \qwenLogo Qwen2.5-7B & 93.52 & 90.04 & 91.78 \\
        \qwenLogo Qwen3-4B & 92.91 & 87.98 & 90.45 \\
        \graniteLogo Granite3.3-8B & 92.94 & 89.20 & 91.07 \\
        \graniteLogo Granite4-3B & 92.76 & 88.63 & 90.70 \\
        \geminiLogo Gemma2-9B & 93.55 & 90.29 & 91.92 \\
        \geminiLogo Gemma3-4B & 93.17 & 88.49 & 90.83 \\
        \hdashline
        \rowcolor{myrowblue}
        \textbf{Average} & 93.25 & 89.10 & \flatAvg{91.17} \\
        \midrule
        \addlinespace[1pt]
        \rowcolor{gray!20}
        \multicolumn{4}{c}{\textit{Open-Source LLMs (Offset-based JSON)}} \\
        \llamaLogo LLaMA3.1-8B & 63.22 & 53.31 & 58.27 \\
        \llamaLogo LLaMA3.2-3B & 56.13 & 31.20 & 43.67 \\
        \qwenLogo Qwen2.5-7B & 60.84 & 50.92 & 55.88 \\
        \qwenLogo Qwen3-4B & 61.18 & 38.34 & 49.76 \\
        \graniteLogo Granite3.3-8B & 64.79 & 44.08 & 54.44 \\
        \graniteLogo Granite4-3B & 61.40 & 38.91 & 50.16 \\
        \geminiLogo Gemma2-9B & 65.53 & 61.45 & 63.49 \\
        \geminiLogo Gemma3-4B & 59.85 & 51.35 & 55.60 \\
        \hdashline
        \rowcolor{myrowblue}
        \textbf{Average} & 61.62 & 46.20 & \flatAvg{53.91} \\
        \bottomrule
    \end{tabularx}
    \caption{Results of pre-trained models, GPT-3, and open-source LLMs on flat NER datasets. The best results are shown in \best{bold} and the second-best results are shown in \secondbest{underlined}. Averages are highlighted in \nestedAvg{blue}.}
    \label{table:flat_ner_results}
\end{table}

\noindent\textbf{Implementation Details.} We fine-tuned all models using the LLaMA-Factory~\cite{zheng-etal-2024-llamafactory} with LoRA~\cite{hu2022lora}. We compare LoRA and full fine-tuning in Appendix~\ref{app:lora_full}, showing that LoRA achieves competitive performance with fewer parameters. We set the LoRA rank $r$ to 256 and the scaling factor $\alpha$ to 512, with adapters inserted into all linear modules. The models were trained with a total batch size of 8 and a learning rate of 2e-5, utilizing a cosine scheduler and a warmup ratio of 0.01. The maximum input length was set to 2048, and all models were trained for two epochs.

\subsection{Results}
The results for flat and nested NER are shown in Table~\ref{table:flat_ner_results} and Table~\ref{table:nested_ner_results}, respectively. We report single-run Micro-F1 scores in the main text and provide the complete Precision, Recall, and F1 results in Appendix~\ref{app:full_prf_results}. The main findings are as follows: 

\textbf{The output format has a decisive impact on the performance of generative NER.} Among the five output formats, the Inline Bracketed and Inline XML formats achieve the best average performance across all LLMs, with average F1 scores of 91.91/82.85 and 91.71/82.67 on flat and nested NER. Both inline formats consistently outperform the JSON-based formats (90.78/81.69, 91.17/80.07 and 53.91/34.53). These results indicate that clear, structured, natural-language-like inline outputs are more conducive to helping LLMs understand and generate accurate entity information, whereas formats requiring concluding the whole information all at once and precise character-offset calculations are poorly suited to the text-generation nature of LLMs, leading to a sharp decline in performance.

\begin{table}[t!]
    \scriptsize
    \centering
    \setlength{\tabcolsep}{3pt}
    \renewcommand{\arraystretch}{1.05}
    \begin{tabularx}{1\columnwidth}{@{}>{\raggedright\arraybackslash}p{0.46\columnwidth}>{\centering\arraybackslash}X>{\centering\arraybackslash}X>{\centering\arraybackslash}X@{}}
        \toprule
        \textbf{Models} & \textbf{ACE2005} & \textbf{GENIA} & \textbf{Average} \\
        \midrule
        \addlinespace[1pt]
        \rowcolor{gray!20}
        \multicolumn{4}{c}{\textit{Pre-Trained NER Models}} \\
        BERT-MRC~\cite{li-etal-2020-unified} & 86.88 & \best{83.75} & 85.32 \\
        BINDER~\cite{zhang2023optimizing} & \best{90.00} & \secondbest{80.80} & 85.40 \\
        W2NER~\cite{Li_Fei_Liu_Wu_Zhang_Teng_Ji_Li_2022} & 86.79 & 80.32 & 83.56 \\
        Biaffine+CNN~\cite{yan-etal-2023-embarrassingly} & 87.25 & 80.33 & 83.79 \\
        \midrule
        \addlinespace[1pt]
        \rowcolor{gray!20}
        \multicolumn{4}{c}{\textit{GPT-NER}~\cite{wang-etal-2025-gpt}} \\
        \chatgptLogo GPT-3 (\textit{ZS}) & 72.96 & 64.06 & 68.51 \\
        \chatgptLogo GPT-3 (\textit{ZS w/ Ver.}) & 73.46 & 64.29 & 68.88 \\
        \chatgptLogo GPT-3 (\textit{FS w/ Ver.}) & 73.59 & 64.42 & 69.01 \\
        \midrule
        \addlinespace[1pt]
        \rowcolor{gray!20}
        \multicolumn{4}{c}{\textit{Open-Source LLMs (Inline Bracketed)}} \\
        \llamaLogo LLaMA3.1-8B & 87.51 & 79.18 & 83.35 \\
        \llamaLogo LLaMA3.2-3B & 85.82 & 79.15 & 82.49 \\
        \qwenLogo Qwen2.5-7B & 86.70 & 78.67 & 82.69 \\
        \qwenLogo Qwen3-4B & 86.48 & 79.16 & 82.82 \\
        \graniteLogo Granite3.3-8B & 86.81 & 78.87 & 82.84 \\
        \graniteLogo Granite4-3B & 86.78 & 78.42 & 82.60 \\
        \geminiLogo Gemma2-9B & 87.44 & 79.73 & 83.59 \\
        \geminiLogo Gemma3-4B & 86.20 & 78.64 & 82.42 \\
        \hdashline
        \rowcolor{myrowblue}
        \textbf{Average} & 86.72 & 78.98 & \nestedAvg{82.85} \\
        \midrule
        \addlinespace[1pt]
        \rowcolor{gray!20}
        \multicolumn{4}{c}{\textit{Open-Source LLMs (Inline XML)}} \\
        \llamaLogo LLaMA3.1-8B & 86.13 & 78.41 & 82.27 \\
        \llamaLogo LLaMA3.2-3B & 86.26 & 78.22 & 82.24 \\
        \qwenLogo Qwen2.5-7B & 85.97 & 79.27 & 82.62 \\
        \qwenLogo Qwen3-4B & 85.66 & 79.60 & 82.63 \\
        \graniteLogo Granite3.3-8B & 87.27 & 79.09 & 83.18 \\
        \graniteLogo Granite4-3B & 86.84 & 78.61 & 82.73 \\
        \geminiLogo Gemma2-9B & \secondbest{88.02} & 79.32 & 83.67 \\
        \geminiLogo Gemma3-4B & 85.62 & 78.46 & 82.04 \\
        \hdashline
        \rowcolor{myrowblue}
        \textbf{Average} & 86.47 & 78.87 & \nestedAvg{82.67} \\
        \midrule
        \addlinespace[1pt]
        \rowcolor{gray!20}
        \multicolumn{4}{c}{\textit{Open-Source LLMs (Category-grouped JSON)}} \\
        \llamaLogo LLaMA3.1-8B & 86.13 & 78.29 & 82.21 \\
        \llamaLogo LLaMA3.2-3B & 83.69 & 77.99 & 80.84 \\
        \qwenLogo Qwen2.5-7B & 84.84 & 78.15 & 81.50 \\
        \qwenLogo Qwen3-4B & 84.35 & 78.48 & 81.42 \\
        \graniteLogo Granite3.3-8B & 86.23 & 78.47 & 82.35 \\
        \graniteLogo Granite4-3B & 84.77 & 78.38 & 81.58 \\
        \geminiLogo Gemma2-9B & 86.64 & 78.93 & 82.79 \\
        \geminiLogo Gemma3-4B & 84.51 & 77.18 & 80.85 \\
        \hdashline
        \rowcolor{myrowblue}
        \textbf{Average} & 85.14 & 78.23 & \nestedAvg{81.69} \\
        \midrule
        \addlinespace[1pt]
        \rowcolor{gray!20}
        \multicolumn{4}{c}{\textit{Open-Source LLMs (Occurrence-based JSON)}} \\
        \llamaLogo LLaMA3.1-8B & 83.41 & 78.66 & 81.04 \\
        \llamaLogo LLaMA3.2-3B & 80.93 & 77.41 & 79.17 \\
        \qwenLogo Qwen2.5-7B & 81.69 & 77.58 & 79.64 \\
        \qwenLogo Qwen3-4B & 81.58 & 78.47 & 80.03 \\
        \graniteLogo Granite3.3-8B & 83.02 & 78.27 & 80.65 \\
        \graniteLogo Granite4-3B & 81.88 & 77.25 & 79.57 \\
        \geminiLogo Gemma2-9B & 84.29 & 79.12 & 81.71 \\
        \geminiLogo Gemma3-4B & 81.58 & 76.03 & 78.81 \\
        \hdashline
        \rowcolor{myrowblue}
        \textbf{Average} & 82.30 & 77.85 & \nestedAvg{80.07} \\
        \midrule
        \addlinespace[1pt]
        \rowcolor{gray!20}
        \multicolumn{4}{c}{\textit{Open-Source LLMs (Offset-based JSON)}} \\
        \llamaLogo LLaMA3.1-8B & 35.85 & 34.62 & 35.24 \\
        \llamaLogo LLaMA3.2-3B & 27.31 & 25.79 & 26.55 \\
        \qwenLogo Qwen2.5-7B & 32.59 & 32.02 & 32.31 \\
        \qwenLogo Qwen3-4B & 35.41 & 36.81 & 36.11 \\
        \graniteLogo Granite3.3-8B & 36.90 & 39.54 & 38.22 \\
        \graniteLogo Granite4-3B & 33.30 & 36.80 & 35.05 \\
        \geminiLogo Gemma2-9B & 40.80 & 41.15 & 40.98 \\
        \geminiLogo Gemma3-4B & 31.60 & 32.00 & 31.80 \\
        \hdashline
        \rowcolor{myrowblue}
        \textbf{Average} & 34.22 & 34.84 & \nestedAvg{34.53} \\
        \bottomrule
    \end{tabularx}
    \caption{Results of pre-trained models, GPT-3, and open-source LLMs on nested NER datasets.}
    \label{table:nested_ner_results}
\end{table}

\begin{table*}[t]
\scalebox{0.87}{
\begin{tabular}{l} 
\toprule
\addlinespace[1pt]
\rowcolor{gray!20}
\multicolumn{1}{c}{\textit{Instruction I. Symbol-based Label with Explanations} (CoNLL2003-SE)}\\
\textbf{Instruction}: Your task is to do sequence labeling with labels A, B, C, and D. \\
 A: A collective entity such as a company, institution, brand, political or governmental body\dots (Explanation of \textsc{ORG}) \\
 B: A named individual, including humans, animals, fictional characters, and their aliases.\quad\quad (Explanation of \textsc{PER}) \\
 C: A geographical or spatial entity, including natural features, built structures, regions\dots\quad\quad (Explanation of \textsc{LOC}) \\
 D: Named entities that are not persons, organizations, or locations, including derived\dots\quad\quad (Explanation of \textsc{MISC})\\

\midrule
\addlinespace[1pt]
\rowcolor{gray!20}
\multicolumn{1}{c}{\textit{Instruction II. Symbol-based Label Only} (CoNLL2003-SO)}\\
\textbf{Instruction}: Your task is to do sequence labeling with labels A, B, C, and D.\\

\midrule
\addlinespace[1pt]
\rowcolor{gray!20}
\multicolumn{1}{c}{\textit{Input Sentence and Output with Symbol-based Labels Appended to the Instructions}}\\
\textbf{Input Sentence}: Havel praises Czech native Albright as friend.\\
\textbf{Output}: \textbf{<B>}Havel\textbf{</B>} praises \textbf{<D>}Czech\textbf{</D>} native \textbf{<B>}Albright\textbf{</B>} as friend.\\
\bottomrule
\end{tabular}}
\caption{Examples of two variants of CoNLL2003 with symbol-based labels. The first contains explanations for each symbolic label (CoNLL2003-SE), while the second does not offer explanations in each sample (CoNLL2003-SO).}
\label{table:memorization}
\end{table*}

\begin{figure*}[t!]
    \centering
    \captionsetup[subfigure]{skip=0.1pt}
    \begin{subfigure}[b]{0.48\textwidth}
        \centering
        \includegraphics[width=0.9\textwidth]{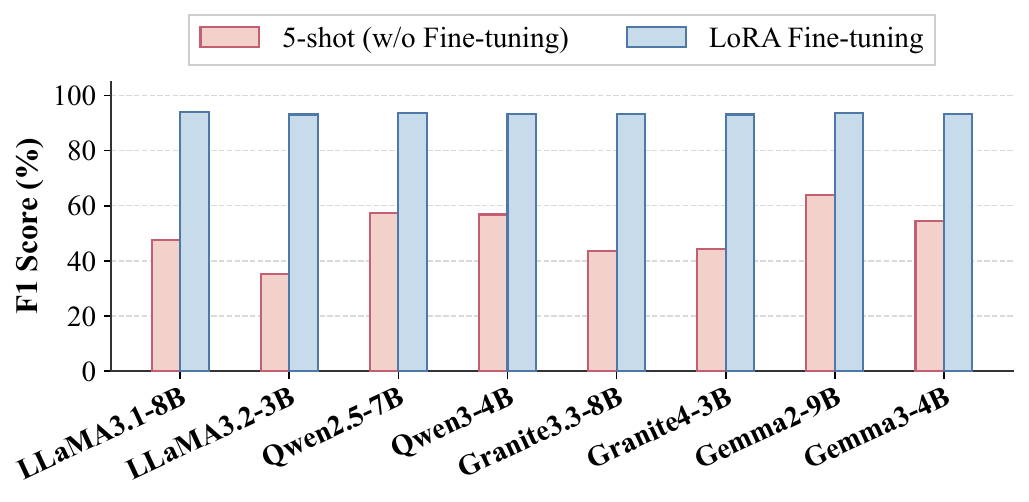}
        \caption{}
        \label{fig:memorization_bar}
    \end{subfigure}
    \hfill
    \begin{subfigure}[b]{0.48\textwidth}
        \centering
        \includegraphics[width=0.9\textwidth]{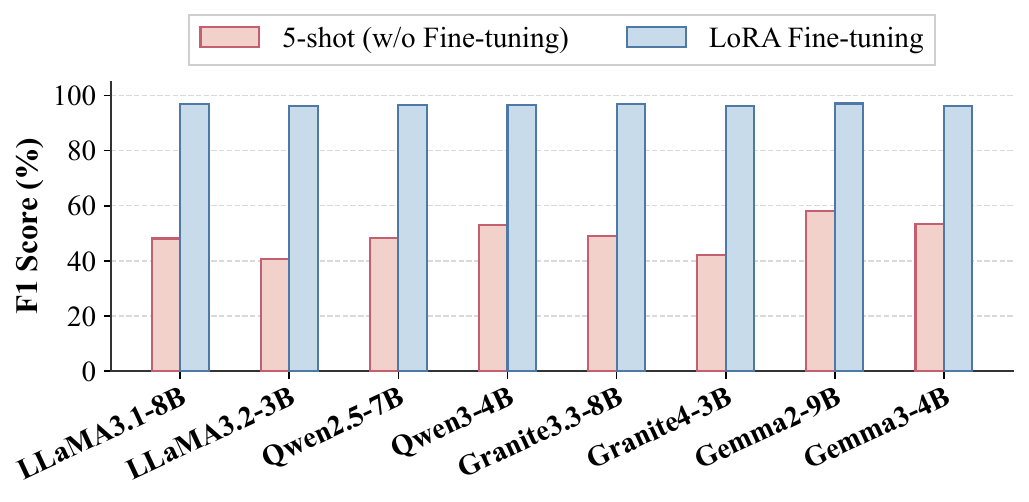}
        \caption{}
        \label{fig:memorization_bar_cleanconll}
    \end{subfigure}

    \vspace{2mm}

    \begin{subfigure}[b]{0.48\textwidth}
        \centering
        \includegraphics[width=\textwidth]{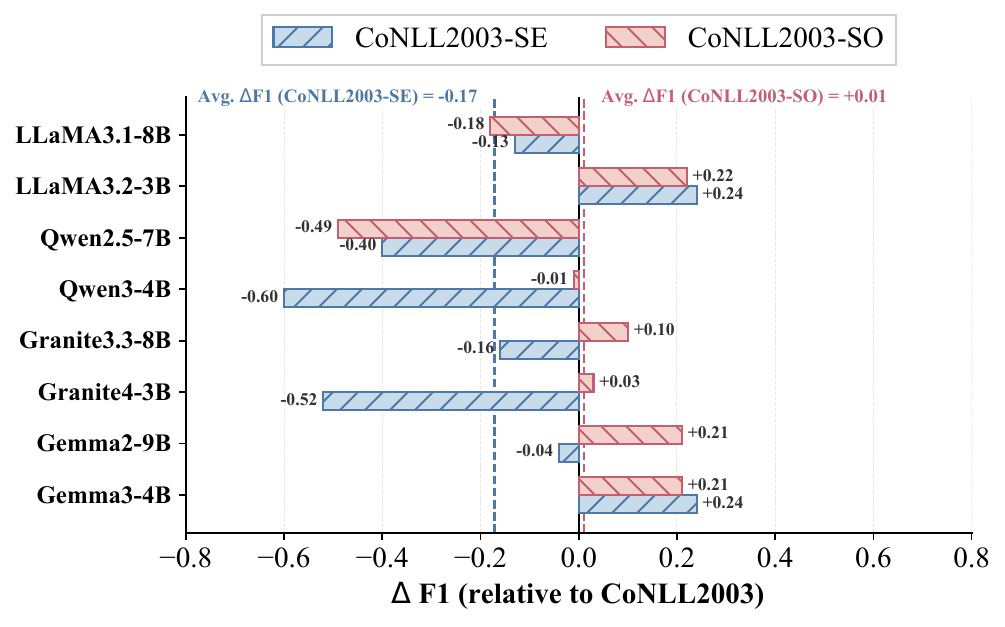}
        \caption{}
        \label{fig:memorization_divergence}
    \end{subfigure}
    \hfill
    \begin{subfigure}[b]{0.48\textwidth}
        \centering
        \includegraphics[width=\textwidth]{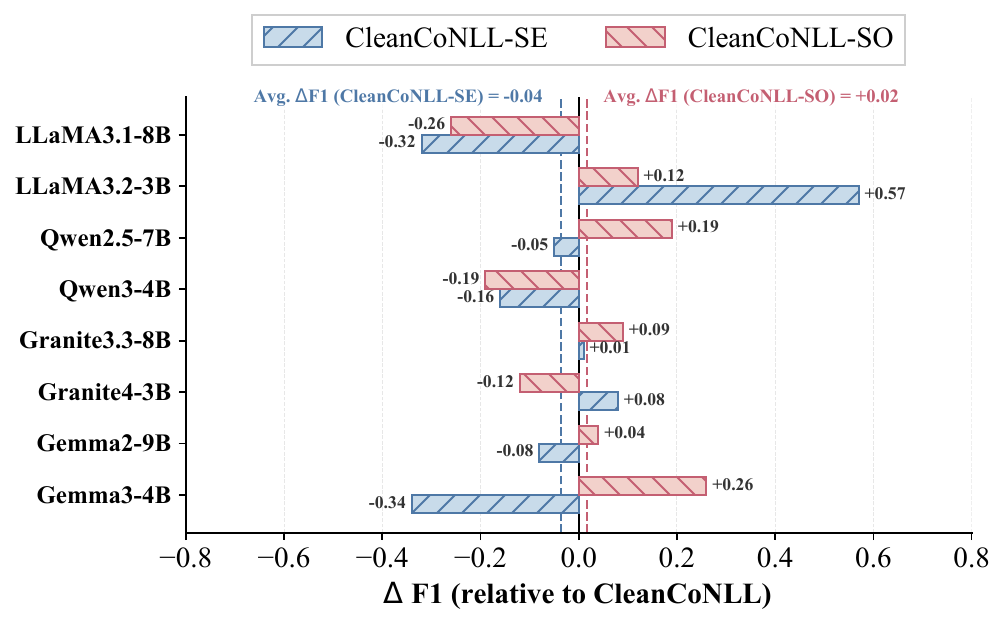}
        \caption{}
        \label{fig:memorization_divergence_cleanconll}
    \end{subfigure}
    \caption{Investigation of entity-label memorization. (a) and (b) compare few-shot prompting with LoRA SFT on CoNLL2003 (left) and CleanCoNLL (right). (c) and (d) report F1 changes under two variants (SE: with explanations; SO: symbol only) relative to the original dataset.}
    \label{fig:results_memorization}
\end{figure*}

\textbf{Well-designed output formats enable LLMs to match or surpass some traditional fine-tuning methods and closed-source models.} With the Inline Bracketed and XML format, LLaMA3.1-8B achieves 93.85 and 93.83 on CoNLL2003, surpassing all pre-trained NER baselines. On ACE2005, Gemma2-9B achieves 88.02, surpassing BERT-MRC (86.88) and Biaffine+CNN (87.25). Both inline formats also substantially outperform in-context learning by closed-source GPT-3 models. This demonstrates that with properly constructed prompts, generative NER with open-source LLMs can achieve highly competitive performance, leveraging powerful native language understanding capabilities. Moreover, as shown in Section~\ref{subsec:general_capability}, after instruction tuning for NER, the model's general capability is preserved, with performance on some tasks remaining unaffected or even improved.

\textbf{Compared with the general domain, LLMs lag behind traditional specialized models in low-resource domains.} Although LLMs achieve competitive results on general-domain datasets, their performance is less competitive on GENIA in the biomedical domain. The best LLM result is 4.02 F1 points lower than the best specialized model BERT-MRC and falls below other traditional baselines. This indicates that due to a lack of prior knowledge in low-resource domains, LLMs' capabilities there are significantly weaker than in the general domains. This highlights the generalization limitations of LLMs in low-resource domains, where their performance is inferior to that in general domains. This gap is again evidenced in NER tasks. Despite LLMs being trained on the same data, their deficiency in low-resource domain knowledge prevents them from surpassing traditional models that have undergone targeted training.

\section{Entity Recognition or Memorization?}
LLMs are trained on extensive datasets during the pre-training process. In this section, we investigate whether LLMs genuinely possess entity \textit{recognition} capabilities or just rely on entity-label \textit{memorization} when performing generative NER tasks.

\begin{table*}[t!]
\centering
\small
\renewcommand{\arraystretch}{1.4}
\begin{tabularx}{\textwidth}{ll X}
\toprule
&\textbf{Error Types} & \textbf{Examples of Incorrectly Labeled Output} \\ \midrule

\multirow{2}{*}{\rotatebox{90}{\textbf{Type}}} 
&\textit{OOD Types}           & \underline{[Inter]$_{\textsc {ORG}}$} will be without suspended \underline{[French]$_{\textsc {MISC}}$} defender \textcolor{blue}{\textbf{[Joceyln Angloma]$_{\textsc {PLAYER}}$}} \\
&\textit{Wrong Types}         & \textcolor{blue}{\textbf{[Inter]$_{\textsc {LOC}}$}} will be without suspended \underline{[French]$_{\textsc {MISC}}$} defender \underline{[Joceyln Angloma]$_{\textsc {PER}}$} \\ 
\midrule
\multirow{3}{*}{\rotatebox{90}{\textbf{Boundary}}} 
&\textit{Contain Gold }       & \underline{[Inter]$_{\textsc {ORG}}$} will be without suspended \underline{[French]$_{\textsc {MISC}}$} \textcolor{blue}{\textbf{[defender Joceyln Angloma]$_{\textsc {PER}}$}} \\
&\textit{Contained by Gold}   & \underline{[Inter]$_{\textsc {ORG}}$} will be without suspended \underline{[French]$_{\textsc {MISC}}$} defender Joceyln \textcolor{blue}{\textbf{[Angloma]$_{\textsc {PER}}$}} \\
&\textit{Overlap with Gold}   & \underline{[Inter]$_{\textsc {ORG}}$} will be without suspended French \textcolor{blue}{\textbf{[defender Joceyln]$_{\textsc {PER}}$}} Angloma \\ 
\midrule
\multirow{3}{*}{\rotatebox{90}{\textbf{Others}}} 
&\textit{Completely-O}        & \underline{[Inter]$_{\textsc {ORG}}$} will be without \textcolor{blue}{\textbf{[suspended]$_{\textsc {MISC}}$}} \underline{[French]$_{\textsc {MISC}}$} defender \underline{[Joceyln Angloma]$_{\textsc {PER}}$} \\
&\textit{OOD Mentions}        & \underline{[Inter]$_{\textsc {ORG}}$} will be without suspended \underline{[French]$_{\textsc {MISC}}$} defender \underline{[Joceyln Angloma]$_{\textsc {PER}}$} \textcolor{blue}{\textbf{[Milan]$_{\textsc {ORG}}$}} \\
&\textit{Omitted Mentions}    & \underline{[Inter]$_{\textsc {ORG}}$} will be without suspended \textcolor{blue}{\textbf{French}} defender \underline{[Joceyln Angloma]$_{\textsc {PER}}$} \\
\bottomrule
\end{tabularx}
\caption{Error types based on the sentence: ``{[Inter]$_{\textsc {ORG}}$} will be without suspended {[French]$_{\textsc {MISC}}$} defender {[Joceyln Angloma]$_{\textsc {PER}}$}.''  \underline{Underlined entities} denote correct predictions, while \textcolor{blue}{\textbf{blue bold}} spans highlight specific errors.}
\label{errortypes}
\end{table*}

\begin{figure*}[t!]
    \centering
    \begin{subfigure}[b]{0.45\textwidth}
        \centering
        \caption{BERT-MRC}
        \label{fig:error_distribution_bert}
        \includegraphics[width=\textwidth]{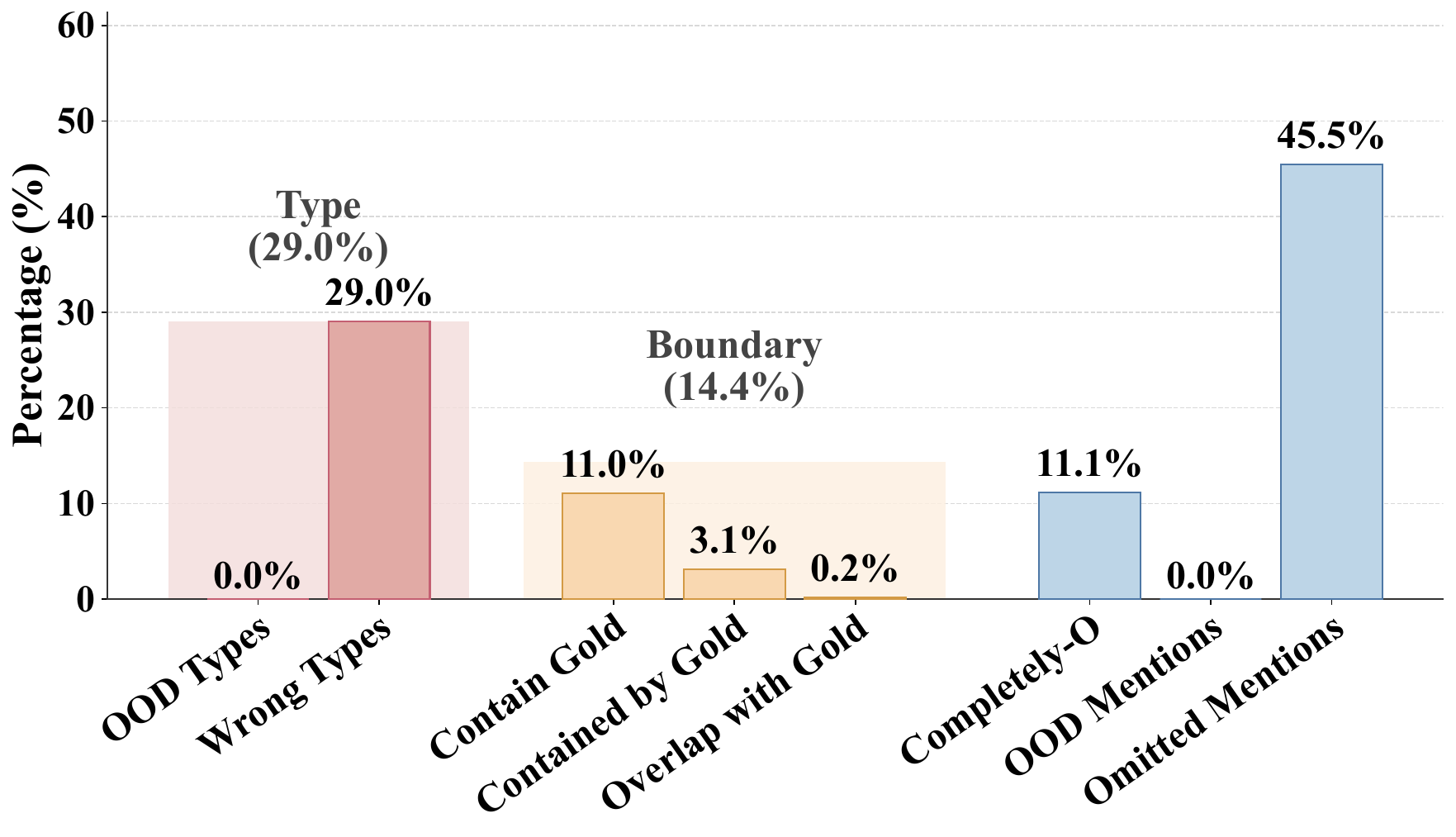}
    \end{subfigure}
    \hfill
    \begin{subfigure}[b]{0.45\textwidth}
        \centering
        \caption{LLaMA3.1-8B}
        \label{fig:error_distribution_llama}
        \includegraphics[width=\textwidth]{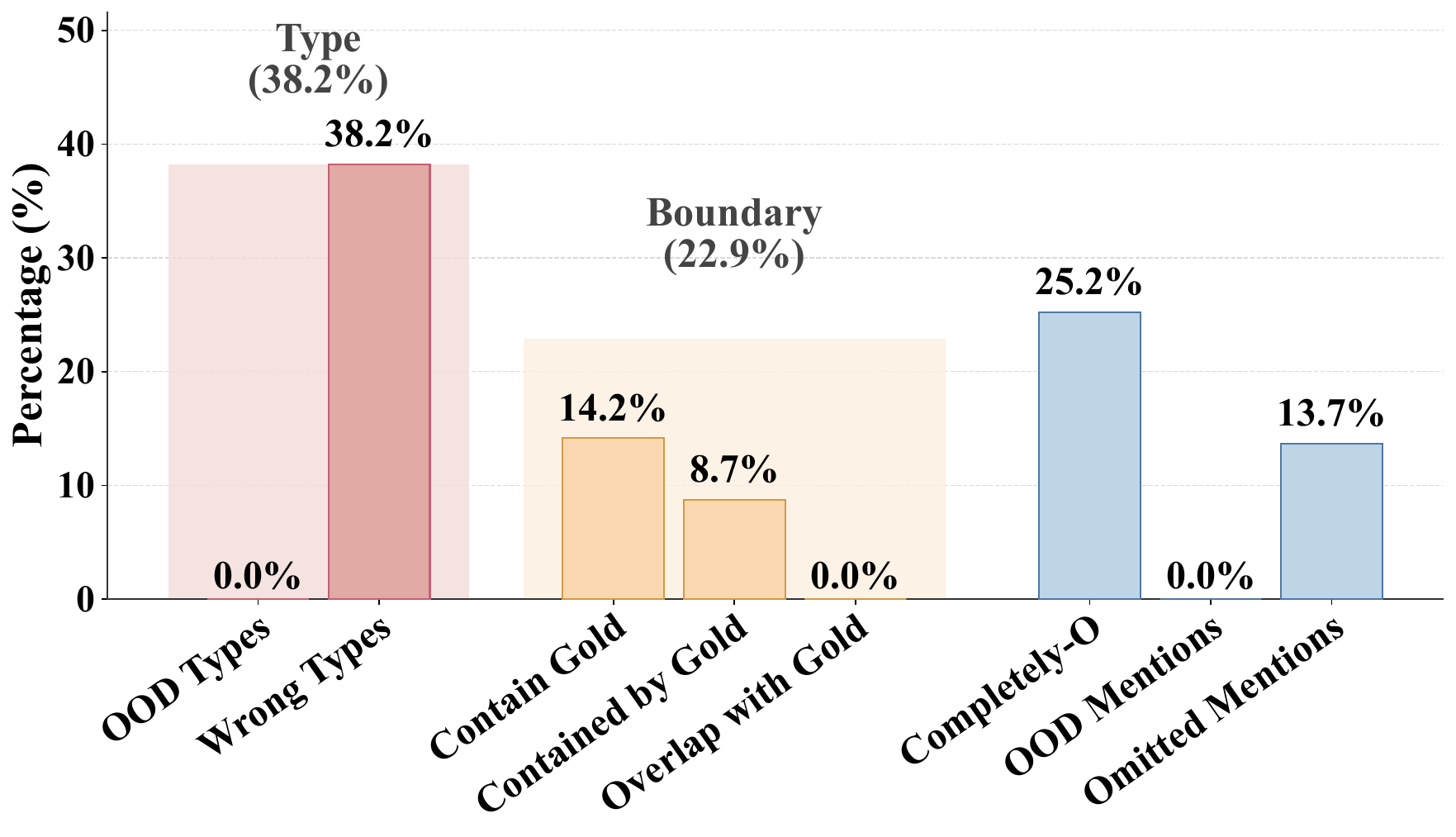}
    \end{subfigure}
    \caption{Distribution of error types for BERT-MRC and LLaMA3.1-8B on CoNLL2003.}
    \label{fig:error_distribution}
\end{figure*}

\noindent\textbf{Settings.} We conduct two experiments on both CoNLL2003 and CleanCoNLL~\cite{rucker-akbik-2023-cleanconll} with the Inline XML format. CleanCoNLL is a nearly noise-free revision of CoNLL2003 with corrected annotation errors. Because it must be reconstructed locally from the original corpus using the released correction scripts, its fully processed form is unlikely to have appeared in model pre-training data, making it a more reliable dataset for investigating memorization. \textbf{Few-shot prompting without fine-tuning:} We evaluate all eight LLMs under a 5-shot in-context learning setting. If LLMs had already memorized entity-label mappings during pre-training, they should achieve competitive performance without fine-tuning. \textbf{Symbol-based label replacement:} We replace the ground-truth text labels (e.g., \textsc{LOC}, \textsc{ORG}) with arbitrary symbols (e.g., \textsc{A}, \textsc{B}) in each sample. We further distinguish two settings based on whether label meaning explanations are provided. In the symbolic label with explanations (SE) setting, we provide standard entity meaning interpretations for the symbols and denote the transformed datasets as CoNLL2003-SE and CleanCoNLL-SE. In the symbol-only (SO) setting, we omit such interpretations, allowing the model to learn the meaning of each symbol solely from the training examples. The transformed datasets are denoted as CoNLL2003-SO and CleanCoNLL-SO.

\noindent\textbf{Results.} As shown in Figure~\ref{fig:results_memorization}(a) and (b), all LLMs achieve moderate F1 scores under 5-shot prompting on both CoNLL2003 and CleanCoNLL, far below fine-tuning performance. This large gap indicates that even if LLMs were exposed to the sentences during pre-training, their generative NER ability remains limited. The NER capability is mainly acquired through instruction tuning.

From the symbol-based label experiment in Figure~\ref{fig:results_memorization}(c) and (d), we find that: 1) LLMs exhibit only marginal F1 changes on both datasets (average $\Delta$F1 of $-$0.17 on CoNLL2003-SE and $+$0.01 on CoNLL2003-SO; $-$0.04 on CleanCoNLL-SE and $+$0.02 on CleanCoNLL-SO); 2) The symbol-only setting even yields slight average increases on both datasets. This may be because symbol explanations introduce additional context that confuses the model, whereas pure symbolic labels reduce inference-time ambiguity.

Overall, we reach the same conclusion on both datasets: \textbf{LLMs do not rely on entity-label memorization in generative NER tasks, but rather demonstrate a considerable capability in entity learning}, thereby effectively accomplishing entity recognition tasks.

\begin{table*}[t!]
\centering
\small
\renewcommand{\arraystretch}{1}
\setlength{\tabcolsep}{6pt}
\renewcommand{\tabularxcolumn}[1]{m{#1}} 

\begin{tabularx}{\linewidth}{l >{\raggedright\arraybackslash}X c c c}
\toprule
\textbf{Error Types} & \textbf{Input Sentence} & \textbf{Ground Truth} & \textbf{BERT-MRC} & \textbf{LLaMA3.1-8B} \\
\midrule

\textit{Omitted Mentions} 
& \mbox{Italy recalled \underline{Marcello Cuttitta}.} 
& [\textbf{M. Cuttitta}]$_{\textsc{PER}}$ 
& \begin{tabular}{@{}c@{}} 
    \textcolor{blue}{\textbf{O} (Missed)} \\ 
    \small (45.5\%) 
  \end{tabular}
& \begin{tabular}{@{}c@{}} 
    [\textbf{M. Cuttitta}]$_{\textsc{PER}}$ \\ 
    \small (13.7\%) 
  \end{tabular} \\
\midrule
\textit{Wrong Types} 
& In congruence with futures, \underline{Permian} \newline cash prices for the weekend fell more...
& [\textbf{Permian}]$_{\textsc{MISC}}$ 
& \begin{tabular}{@{}c@{}} 
    [\textbf{Permian}]$_{\textsc{MISC}}$ \\ 
    \small (29.0\%) 
  \end{tabular}
& \begin{tabular}{@{}c@{}} 
    \textcolor{blue}{[\textbf{Permian}]$_{\textsc{LOC}}$} \\ 
    \small (38.2\%) 
  \end{tabular} \\
\midrule
\textit{Completely-O} 
& I've also got a contract to play for New\newline South Wales in the \underline{Super 12} next year. 
& \textbf{O} 
& \begin{tabular}{@{}c@{}} 
    \textbf{O} \\ 
    \small (11.1\%) 
  \end{tabular}
& \begin{tabular}{@{}c@{}} 
    \textcolor{blue}{[\textbf{Super 12}]$_{\textsc{MISC}}$} \\ 
    \small (25.2\%) 
  \end{tabular} \\

\bottomrule
\end{tabularx}
\caption{Case study of errors from BERT-MRC and LLaMA3.1-8B. Text \textcolor{blue}{in blue} indicates incorrect predictions. The number in parentheses indicates the share of each error type among all errors.}
\label{tab:case_study}
\end{table*}

\begin{table*}[t!]
\centering
\resizebox{\textwidth}{!}{
\begin{tabular}{lcccccc} 
\toprule
\multirow{2}{*}{\textbf{Models}} & \textbf{TruthfulQA} & \textbf{MMLU} & \textbf{GSM8K} & \textbf{HellaSwag} & \textbf{DROP} & \multirow{2}{*}{\textbf{Avg.}} \\
 & MC1 & Acc & EM & Acc\_norm & F1 &  \\
\midrule
\textbf{LLaMA3.1-8B} & 40.15 & 68.76 & 84.69 & 72.53 & 33.20 & 59.87 \\
\textit{w/ NER LoRA Fine-tuning (XML)} & 36.96~\textcolor{magenta}{($\downarrow$3.19)} & 67.08~\textcolor{magenta}{($\downarrow$1.68)} & 80.36~\textcolor{magenta}{($\downarrow$4.33)} & 70.66~\textcolor{magenta}{($\downarrow$1.87)} & 58.70~\textcolor{teal}{\textbf{($\uparrow$25.50)}} & 62.75~\textcolor{teal}{($\uparrow$2.88)} \\
\midrule
\textbf{Qwen2.5-7B} & 47.49 & 73.42 & 76.57 & 65.43 & 15.74 & 55.73 \\
\textit{w/ NER LoRA Fine-tuning (XML)} & 45.04~\textcolor{magenta}{($\downarrow$2.45)} & 72.75~\textcolor{magenta}{($\downarrow$0.67)} & 79.61~\textcolor{teal}{($\uparrow$3.04)} & 68.47~\textcolor{teal}{($\uparrow$3.04)} & 61.06~\textcolor{teal}{\textbf{($\uparrow$45.32)}} & 65.39~\textcolor{teal}{($\uparrow$9.66)} \\
\bottomrule
\end{tabular}
}
\caption{Performance comparison of LLMs on benchmarks before and after generative NER fine-tuning.}
\label{tab:general_capability}
\end{table*}

\section{Error Analysis on Generative NER}
To understand what types of errors LLMs are prone to make in NER and their specific gaps compared to traditional models, we conduct a statistical analysis of error samples. We follow \citet{xie-etal-2023-empirical} and categorize errors into eight types, as shown in Table~\ref{errortypes}. Detailed definitions are in Appendix~\ref{app:error_definitions}. 

\noindent\textbf{Statistical Results.}  We present the error distribution statistics for the representative pre-trained model BERT-MRC, and the fine-tuned LLaMA3.1-8B with Inline
Bracketed format in Figure~\ref{fig:error_distribution}. 

Overall, we observe clear differences in error patterns between the two paradigms. BERT-MRC's errors are dominated by \textit{Omitted Mentions} (45.5\%), indicating a conservative prediction strategy that tends to miss entities. 
In contrast, LLaMA3.1-8B exhibits a substantially lower omission rate, reflecting a more aggressive extraction behavior enabled by its extensive pre-training knowledge. However, this higher sensitivity to potential entities leads to a trade-off: the model is prone to over-extraction, resulting in a significantly higher rate of \textit{Completely-O} errors than BERT-MRC.

The primary error for LLMs is \textit{Wrong Types}, comprising 38.2\% of all errors. This shows that while generative models correctly identify entity boundaries, they often misclassify types. This likely arises from a conflict between the model's strong pre-trained prior knowledge and the specific dataset annotation guidelines. Notably, both models have 0\% in \textit{OOD Types} and \textit{OOD Mentions}, confirming their adherence to the predefined label set and that all predictions are grounded in the input text.

\noindent\textbf{Case Study.} To intuitively understand the statistical differences discussed above, we provide typical error examples in Table~\ref{tab:case_study}. 

The first example illustrates a common omission error made by traditional NER models. BERT-MRC omits ``\textit{Marcello Cuttitta}'' due to limited pre-training exposure, whereas the LLM correctly extracts it, leveraging extensive world knowledge to recognize unseen entities.

The second example highlights a typical type misclassification made by LLMs. In the phrase \textit{``Permian cash price''}, the gold standard labels \textit{``Permian''} as \textsc{MISC}, following the CoNLL2003 guideline that adjectival or derived forms from named entities (e.g., locations, organizations, or persons) should be categorized as miscellaneous. LLaMA3.1-8B predicts \textit{``Permian''} as \textsc{LOC}, which can be attributed to prior knowledge during large-scale pre-training that associates \textit{``Permian''} with a geographical region, instead of following the CoNLL2003 annotation guidelines.

The third example demonstrates an over-extraction behavior of LLMs. In this sentence, the gold annotation does not label \textit{``Super 12''}. However, LLaMA3.1-8B incorrectly labels it as \textsc{MISC}. According to the CoNLL2003 annotation guidelines, the \textsc{MISC} category encompasses a broad and diverse range of entities. This lack of precise semantic boundaries creates ambiguity, making LLMs prone to over-predicting entities.

\section{Preservation of General Capability}
\label{subsec:general_capability}
While traditional pre-trained models become single-purpose specialists after task-specific fine-tuning, LLMs retain their inherent multi-task capabilities as general generative models. To examine the impact of generative NER fine-tuning on these general capabilities, we conduct an evaluation across a range of tasks, including reading comprehension (DROP;~\citealp{dua-etal-2019-drop}), commonsense reasoning (HellaSwag;~\citealp{zellers-etal-2019-hellaswag}), multitask language understanding (MMLU;~\citealp{hendrycks2021measuring}), mathematical reasoning (GSM8K;~\citealp{cobbe2021trainingverifierssolvemath}), and truthfulness (TruthfulQA;~\citealp{lin-etal-2022-truthfulqa}). We use the lm-evaluation-harness framework~\cite{eval-harness} for comparison.

The results are shown in Table~\ref{tab:general_capability}. We find that the fine-tuned LLMs performed comparably to the original model across most of the tasks, with only minor fluctuations ranging from a 4.33\% decrease to a 3.04\% increase. Notably, on the reading comprehension benchmark DROP, the fine-tuned model \textbf{achieves a substantial performance gain of 25.50 to 45.32 F1 scores}. The improvement could be attributed to the fact that DROP answers are often text spans containing named entities (e.g., persons, locations, organizations). Fine-tuning for NER strengthens the model’s ability to detect these entities, leading to more precise answer extraction. Overall, the results suggest that the core capabilities of LLMs remain largely intact after undergoing lightweight fine-tuning for the NER task.

\section{Conclusion}
We present a comprehensive assessment of generative NER using open-source LLMs. Our experiments show that LLMs, when fine-tuned with parameter-efficient methods and appropriate output formats, can match or even exceed the performance of strong traditional NER models, while maintaining robust multi-task capabilities. Furthermore, we find that LLMs keep genuine learning on NER task instead of memorizing entity-label correlations. These findings confirm the viability of generative NER as a flexible and effective paradigm, highlighting its potential to gradually supplement traditional sequence labeling approaches.

\section*{Limitations}
This study primarily focuses on evaluating the NER capabilities of LLaMA, Qwen, Granite, and Gemma model families under instruction tuning. Future improvements could be pursued along two dimensions: First, at the model level, the investigation could be extended to reasoning models with long chain-of-thought output, which offer the advantage of tracing entity labels through the reasoning chain, thereby enhancing interpretability. The current work emphasizes instruction compliance more than the model’s thinking and reasoning abilities. Second, regarding training methodology, reinforcement learning-based approaches such as PPO, DPO, and GRPO have been shown to surpass traditional supervised fine-tuning in terms of eliciting model capability and generalization, yet were not explored in this work. These limitations represent key directions for our future research.

\section*{Acknowledgments}

This work was supported in part by ICFCRT (W2441020), Guangdong Basic and Applied Basic Research Foundation (2026A1515011358), Youth S\&T Talent Support Programme of Guangdong Provincial Association for Science and Technology (SKXRC2026862), Shenzhen Natural Science Foundation (JCYJ20250604181610014), Intelligent Computing Center of Shenzhen University, and Scientific Development Fund from Guangdong Provincial Key Laboratory of Visual Media and Multidimensional Intelligence.

\bibliography{custom}

@inproceedings{wang-etal-2025-gpt,
    title = "{GPT}-{NER}: Named Entity Recognition via Large Language Models",
    author = "Wang, Shuhe  and
      Sun, Xiaofei  and
      Li, Xiaoya  and
      Ouyang, Rongbin  and
      Wu, Fei  and
      Zhang, Tianwei  and
      Li, Jiwei  and
      Wang, Guoyin  and
      Guo, Chen",
    editor = "Chiruzzo, Luis  and
      Ritter, Alan  and
      Wang, Lu",
    booktitle = "Findings of the Association for Computational Linguistics: NAACL 2025",
    month = apr,
    year = "2025",
    address = "Albuquerque, New Mexico",
    publisher = "Association for Computational Linguistics",
    url = "https://aclanthology.org/2025.findings-naacl.239/",
    doi = "10.18653/v1/2025.findings-naacl.239",
    pages = "4257--4275",
    ISBN = "979-8-89176-195-7"
}

@inproceedings{xie-etal-2023-empirical,
    title = "Empirical Study of Zero-Shot {NER} with {C}hat{GPT}",
    author = "Xie, Tingyu  and
      Li, Qi  and
      Zhang, Jian  and
      Zhang, Yan  and
      Liu, Zuozhu  and
      Wang, Hongwei",
    editor = "Bouamor, Houda  and
      Pino, Juan  and
      Bali, Kalika",
    booktitle = "Proceedings of the 2023 Conference on Empirical Methods in Natural Language Processing",
    month = dec,
    year = "2023",
    address = "Singapore",
    publisher = "Association for Computational Linguistics",
    url = "https://aclanthology.org/2023.emnlp-main.493/",
    doi = "10.18653/v1/2023.emnlp-main.493",
    pages = "7935--7956"
}

@inproceedings{ma-etal-2022-template,
    title = "Template-free Prompt Tuning for Few-shot {NER}",
    author = "Ma, Ruotian  and
      Zhou, Xin  and
      Gui, Tao  and
      Tan, Yiding  and
      Li, Linyang  and
      Zhang, Qi  and
      Huang, Xuanjing",
    editor = "Carpuat, Marine  and
      de Marneffe, Marie-Catherine  and
      Meza Ruiz, Ivan Vladimir",
    booktitle = "Proceedings of the 2022 Conference of the North American Chapter of the Association for Computational Linguistics: Human Language Technologies",
    month = jul,
    year = "2022",
    address = "Seattle, United States",
    publisher = "Association for Computational Linguistics",
    url = "https://aclanthology.org/2022.naacl-main.420/",
    doi = "10.18653/v1/2022.naacl-main.420",
    pages = "5721--5732"
}

@inproceedings{devlin-etal-2019-bert,
    title = "{BERT}: Pre-training of Deep Bidirectional Transformers for Language Understanding",
    author = "Devlin, Jacob  and
      Chang, Ming-Wei  and
      Lee, Kenton  and
      Toutanova, Kristina",
    editor = "Burstein, Jill  and
      Doran, Christy  and
      Solorio, Thamar",
    booktitle = "Proceedings of the 2019 Conference of the North {A}merican Chapter of the Association for Computational Linguistics: Human Language Technologies, Volume 1 (Long and Short Papers)",
    month = jun,
    year = "2019",
    address = "Minneapolis, Minnesota",
    publisher = "Association for Computational Linguistics",
    url = "https://aclanthology.org/N19-1423/",
    doi = "10.18653/v1/N19-1423",
    pages = "4171--4186"
}

@inproceedings{lewis-etal-2020-bart,
    title = "{BART}: Denoising Sequence-to-Sequence Pre-training for Natural Language Generation, Translation, and Comprehension",
    author = "Lewis, Mike  and
      Liu, Yinhan  and
      Goyal, Naman  and
      Ghazvininejad, Marjan  and
      Mohamed, Abdelrahman  and
      Levy, Omer  and
      Stoyanov, Veselin  and
      Zettlemoyer, Luke",
    editor = "Jurafsky, Dan  and
      Chai, Joyce  and
      Schluter, Natalie  and
      Tetreault, Joel",
    booktitle = "Proceedings of the 58th Annual Meeting of the Association for Computational Linguistics",
    month = jul,
    year = "2020",
    address = "Online",
    publisher = "Association for Computational Linguistics",
    url = "https://aclanthology.org/2020.acl-main.703/",
    doi = "10.18653/v1/2020.acl-main.703",
    pages = "7871--7880"
}

@inproceedings{cui-etal-2021-template,
    title = "Template-Based Named Entity Recognition Using {BART}",
    author = "Cui, Leyang  and
      Wu, Yu  and
      Liu, Jian  and
      Yang, Sen  and
      Zhang, Yue",
    editor = "Zong, Chengqing  and
      Xia, Fei  and
      Li, Wenjie  and
      Navigli, Roberto",
    booktitle = "Findings of the Association for Computational Linguistics: ACL-IJCNLP 2021",
    month = aug,
    year = "2021",
    address = "Online",
    publisher = "Association for Computational Linguistics",
    url = "https://aclanthology.org/2021.findings-acl.161/",
    doi = "10.18653/v1/2021.findings-acl.161",
    pages = "1835--1845"
}

@inproceedings{li-etal-2020-unified,
    title = "A Unified {MRC} Framework for Named Entity Recognition",
    author = "Li, Xiaoya  and
      Feng, Jingrong  and
      Meng, Yuxian  and
      Han, Qinghong  and
      Wu, Fei  and
      Li, Jiwei",
    editor = "Jurafsky, Dan  and
      Chai, Joyce  and
      Schluter, Natalie  and
      Tetreault, Joel",
    booktitle = "Proceedings of the 58th Annual Meeting of the Association for Computational Linguistics",
    month = jul,
    year = "2020",
    address = "Online",
    publisher = "Association for Computational Linguistics",
    url = "https://aclanthology.org/2020.acl-main.519/",
    doi = "10.18653/v1/2020.acl-main.519",
    pages = "5849--5859"
}

@inproceedings{kim-etal-2024-exploring,
    title = "Exploring Nested Named Entity Recognition with Large Language Models: Methods, Challenges, and Insights",
    author = "Kim, Hongjin  and
      Kim, Jai-Eun  and
      Kim, Harksoo",
    editor = "Al-Onaizan, Yaser  and
      Bansal, Mohit  and
      Chen, Yun-Nung",
    booktitle = "Proceedings of the 2024 Conference on Empirical Methods in Natural Language Processing",
    month = nov,
    year = "2024",
    address = "Miami, Florida, USA",
    publisher = "Association for Computational Linguistics",
    url = "https://aclanthology.org/2024.emnlp-main.492/",
    doi = "10.18653/v1/2024.emnlp-main.492",
    pages = "8653--8670"
}

@article{wei2023zero,
  title={Zero-Shot Information Extraction via Chatting with ChatGPT},
  author={Wei, Xiang and Cui, Xingyu and Cheng, Ning and Wang, Xiaobin and Zhang, Xin and Huang, Shen and Xie, Pengjun and Xu, Jinan and Chen, Yufeng and Zhang, Meishan and others},
  journal={arXiv preprint arXiv:2302.10205},
url="https://arxiv.org/abs/2302.10205",
  year={2023}
}

@inproceedings{10.1145/3696410.3714923,
author = {Wang, Zihan and Zhao, Ziqi and Lyu, Yougang and Chen, Zhumin and de Rijke, Maarten and Ren, Zhaochun},
title = {A Cooperative Multi-Agent Framework for Zero-Shot Named Entity Recognition},
year = {2025},
isbn = {9798400712746},
publisher = {Association for Computing Machinery},
address = {New York, NY, USA},
url = {https://doi.org/10.1145/3696410.3714923},
doi = {10.1145/3696410.3714923},
booktitle = {Proceedings of the ACM on Web Conference 2025},
pages = {4183–4195},
numpages = {13},
location = {Sydney NSW, Australia},
series = {WWW '25}
}

@inproceedings{ma-hovy-2016-end,
    title = "End-to-end Sequence Labeling via Bi-directional {LSTM}-{CNN}s-{CRF}",
    author = "Ma, Xuezhe  and
      Hovy, Eduard",
    editor = "Erk, Katrin  and
      Smith, Noah A.",
    booktitle = "Proceedings of the 54th Annual Meeting of the Association for Computational Linguistics (Volume 1: Long Papers)",
    month = aug,
    year = "2016",
    address = "Berlin, Germany",
    publisher = "Association for Computational Linguistics",
    url = "https://aclanthology.org/P16-1101/",
    doi = "10.18653/v1/P16-1101",
    pages = "1064--1074"
}

@inproceedings{yang-etal-2018-design,
    title = "Design Challenges and Misconceptions in Neural Sequence Labeling",
    author = "Yang, Jie  and
      Liang, Shuailong  and
      Zhang, Yue",
    editor = "Bender, Emily M.  and
      Derczynski, Leon  and
      Isabelle, Pierre",
    booktitle = "Proceedings of the 27th International Conference on Computational Linguistics",
    month = aug,
    year = "2018",
    address = "Santa Fe, New Mexico, USA",
    publisher = "Association for Computational Linguistics",
    url = "https://aclanthology.org/C18-1327/",
    pages = "3879--3889"
}

@article{Huang_Chen_Huang_Lin_Qin_2026, title={A Reasoning Paradigm for Named Entity Recognition}, volume={40}, url={https://ojs.aaai.org/index.php/AAAI/article/view/40375}, DOI={10.1609/aaai.v40i37.40375}, abstractNote={Generative LLMs typically improve Named Entity Recognition (NER) performance through instruction tuning. They excel at generating entities by semantic pattern matching but lack an explicit, verifiable reasoning mechanism. This &amp;quot;cognitive shortcutting&amp;quot; leads to suboptimal performance and weak generalization, especially in zero-shot and low-resource scenarios where reasoning from limited contextual cues is crucial.
To address this issue, a reasoning framework is proposed for NER, which shifts the extraction paradigm from implicit pattern matching to explicit reasoning. This framework consists of three stages: Chain of Thought (CoT) generation, CoT tuning, and reasoning enhancement. First, a dataset annotated with NER-oriented CoTs is generated, which contain task-relevant reasoning chains. Then, they are used to tune the NER model to generate coherent rationales before deriving the final answer. Finally, a reasoning enhancement stage is implemented to optimize the reasoning process using a comprehensive reward signal. This stage ensures explicit and verifiable extractions.
Experiments show that ReasoningNER demonstrates impressive cognitive ability in the NER task, achieving competitive performance. In zero-shot settings, it achieves SoTA performance, outperforming GPT-4 by 12.3 percentage points on the F1 score. Analytical results demonstrate its great potential to advance research in reasoning-oriented information extraction.}, number={37}, journal={Proceedings of the AAAI Conference on Artificial Intelligence}, author={Huang, Hui and Chen, Yanping and Huang, Ruizhang and Lin, Chuan and Qin, Yongbin}, year={2026}, month={Mar.}, pages={31140–31148} }

@inproceedings{tjong-kim-sang-de-meulder-2003-introduction,
    title = "Introduction to the {C}o{NLL}-2003 Shared Task: Language-Independent Named Entity Recognition",
    author = "Tjong Kim Sang, Erik F.  and
      De Meulder, Fien",
    booktitle = "Proceedings of the Seventh Conference on Natural Language Learning at {HLT}-{NAACL} 2003",
    year = "2003",
    url = "https://aclanthology.org/W03-0419/",
    pages = "142--147"
}

@article{GENIA,
    author = {Kim, J.-D. and Ohta, T. and Tateisi, Y. and Tsujii, J.},
    title = {GENIA corpus—a semantically annotated corpus for bio-textmining},
    journal = {Bioinformatics},
    volume = {19},
    pages = {i180-i182},
    year = {2003},
    month = {07},
    issn = {1367-4803},
    doi = {10.1093/bioinformatics/btg1023},
    url = {https://doi.org/10.1093/bioinformatics/btg1023},
}

@article{grattafiori2024llama,
  title={The llama 3 herd of models},
  author={Grattafiori, Aaron and Dubey, Abhimanyu and Jauhri, Abhinav and Pandey, Abhinav and Kadian, Abhishek and Al-Dahle, Ahmad and Letman, Aiesha and Mathur, Akhil and Schelten, Alan and Vaughan, Alex and others},
  journal={arXiv preprint arXiv:2407.21783},
  url={https://arxiv.org/abs/2407.21783},
  year={2024}
}

@article{yang2025qwen3,
  title={Qwen3 technical report},
  author={Yang, An and Li, Anfeng and Yang, Baosong and Zhang, Beichen and Hui, Binyuan and Zheng, Bo and Yu, Bowen and Gao, Chang and Huang, Chengen and Lv, Chenxu and others},
  journal={arXiv preprint arXiv:2505.09388},
url="https://arxiv.org/abs/2505.09388",
  year={2025}
}

@inproceedings{wang-etal-2023-gnn,
    title = "{GNN}-{SL}: Sequence Labeling Based on Nearest Examples via {GNN}",
    author = "Wang, Shuhe  and
      Meng, Yuxian  and
      Ouyang, Rongbin  and
      Li, Jiwei  and
      Zhang, Tianwei  and
      Lyu, Lingjuan  and
      Wang, Guoyin",
    editor = "Rogers, Anna  and
      Boyd-Graber, Jordan  and
      Okazaki, Naoaki",
    booktitle = "Findings of the Association for Computational Linguistics: ACL 2023",
    month = jul,
    year = "2023",
    address = "Toronto, Canada",
    publisher = "Association for Computational Linguistics",
    url = "https://aclanthology.org/2023.findings-acl.803/",
    doi = "10.18653/v1/2023.findings-acl.803",
    pages = "12679--12692"
}

@inproceedings{wang-etal-2021-automated,
    title = "Automated Concatenation of Embeddings for Structured Prediction",
    author = "Wang, Xinyu  and
      Jiang, Yong  and
      Bach, Nguyen  and
      Wang, Tao  and
      Huang, Zhongqiang  and
      Huang, Fei  and
      Tu, Kewei",
    editor = "Zong, Chengqing  and
      Xia, Fei  and
      Li, Wenjie  and
      Navigli, Roberto",
    booktitle = "Proceedings of the 59th Annual Meeting of the Association for Computational Linguistics and the 11th International Joint Conference on Natural Language Processing (Volume 1: Long Papers)",
    month = aug,
    year = "2021",
    address = "Online",
    publisher = "Association for Computational Linguistics",
    url = "https://aclanthology.org/2021.acl-long.206/",
    doi = "10.18653/v1/2021.acl-long.206",
    pages = "2643--2660"
}

@inproceedings{pradhan-etal-2013-towards,
    title = "Towards Robust Linguistic Analysis using {O}nto{N}otes",
    author = {Pradhan, Sameer  and
      Moschitti, Alessandro  and
      Xue, Nianwen  and
      Ng, Hwee Tou  and
      Bj{\"o}rkelund, Anders  and
      Uryupina, Olga  and
      Zhang, Yuchen  and
      Zhong, Zhi},
    editor = "Hockenmaier, Julia  and
      Riedel, Sebastian",
    booktitle = "Proceedings of the Seventeenth Conference on Computational Natural Language Learning",
    month = aug,
    year = "2013",
    address = "Sofia, Bulgaria",
    publisher = "Association for Computational Linguistics",
    url = "https://aclanthology.org/W13-3516/",
    pages = "143--152"
}

@inproceedings{
zhang2023optimizing,
title={Optimizing Bi-Encoder for Named Entity Recognition via Contrastive Learning},
author={Sheng Zhang and Hao Cheng and Jianfeng Gao and Hoifung Poon},
booktitle={The Eleventh International Conference on Learning Representations },
year={2023},
url={https://openreview.net/forum?id=9EAQVEINuum}
}

@inproceedings{zheng-etal-2024-llamafactory,
    title = "{L}lama{F}actory: Unified Efficient Fine-Tuning of 100+ Language Models",
    author = "Zheng, Yaowei  and
      Zhang, Richong  and
      Zhang, Junhao  and
      Ye, Yanhan  and
      Luo, Zheyan",
    editor = "Cao, Yixin  and
      Feng, Yang  and
      Xiong, Deyi",
    booktitle = "Proceedings of the 62nd Annual Meeting of the Association for Computational Linguistics (Volume 3: System Demonstrations)",
    month = aug,
    year = "2024",
    address = "Bangkok, Thailand",
    publisher = "Association for Computational Linguistics",
    url = "https://aclanthology.org/2024.acl-demos.38/",
    doi = "10.18653/v1/2024.acl-demos.38",
    pages = "400--410"
}

@inproceedings{
hu2022lora,
title={Lo{RA}: Low-Rank Adaptation of Large Language Models},
author={Edward J Hu and yelong shen and Phillip Wallis and Zeyuan Allen-Zhu and Yuanzhi Li and Shean Wang and Lu Wang and Weizhu Chen},
booktitle={International Conference on Learning Representations},
year={2022},
url={https://openreview.net/forum?id=nZeVKeeFYf9}
}

@inproceedings{yan-etal-2023-embarrassingly,
    title = "An Embarrassingly Easy but Strong Baseline for Nested Named Entity Recognition",
    author = "Yan, Hang  and
      Sun, Yu  and
      Li, Xiaonan  and
      Qiu, Xipeng",
    editor = "Rogers, Anna  and
      Boyd-Graber, Jordan  and
      Okazaki, Naoaki",
    booktitle = "Proceedings of the 61st Annual Meeting of the Association for Computational Linguistics (Volume 2: Short Papers)",
    month = jul,
    year = "2023",
    address = "Toronto, Canada",
    publisher = "Association for Computational Linguistics",
    url = "https://aclanthology.org/2023.acl-short.123/",
    doi = "10.18653/v1/2023.acl-short.123",
    pages = "1442--1452"
}

@article{Li_Fei_Liu_Wu_Zhang_Teng_Ji_Li_2022, title={Unified Named Entity Recognition as Word-Word Relation Classification}, volume={36}, url={https://ojs.aaai.org/index.php/AAAI/article/view/21344}, DOI={10.1609/aaai.v36i10.21344}, abstractNote={So far, named entity recognition (NER) has been involved with three major types, including flat, overlapped (aka. nested), and discontinuous NER, which have mostly been studied individually. Recently, a growing interest has been built for unified NER, tackling the above three jobs concurrently with one single model. Current best-performing methods mainly include span-based and sequence-to-sequence models, where unfortunately the former merely focus on boundary identification and the latter may suffer from exposure bias. In this work, we present a novel alternative by modeling the unified NER as word-word relation classification, namely W^2NER. The architecture resolves the kernel bottleneck of unified NER by effectively modeling the neighboring relations between entity words with Next-Neighboring-Word (NNW) and Tail-Head-Word-* (THW-*) relations. Based on the W^2NER scheme we develop a neural framework, in which the unified NER is modeled as a 2D grid of word pairs. We then propose multi-granularity 2D convolutions for better refining the grid representations. Finally, a co-predictor is used to sufficiently reason the word-word relations. We perform extensive experiments on 14 widely-used benchmark datasets for flat, overlapped, and discontinuous NER (8 English and 6 Chinese datasets), where our model beats all the current top-performing baselines, pushing the state-of-the-art performances of unified NER.}, number={10}, journal={Proceedings of the AAAI Conference on Artificial Intelligence}, author={Li, Jingye and Fei, Hao and Liu, Jiang and Wu, Shengqiong and Zhang, Meishan and Teng, Chong and Ji, Donghong and Li, Fei}, year={2022}, month={Jun.}, pages={10965-10973} }

@article{Mu_Ning_Zhao_Zhang_2026, 
  title={A Multi-Agent LLM Framework for Multi-Domain Low-Resource In-Context NER via Knowledge Retrieval, Disambiguation and Reflective Analysis}, 
  volume={40}, 
  url={https://ojs.aaai.org/index.php/AAAI/article/view/40529}, 
  DOI={10.1609/aaai.v40i38.40529}, 
  number={38}, 
  journal={Proceedings of the AAAI Conference on Artificial Intelligence}, 
  author={Mu, Wenxuan and Ning, Jinzhong and Zhao, Di and Zhang, Yijia}, 
  year={2026}, 
  month={Mar.}, 
  pages={32528-32536}
}

@inproceedings{
zhou2024universalner,
title={Universal{NER}: Targeted Distillation from Large Language Models for Open Named Entity Recognition},
author={Wenxuan Zhou and Sheng Zhang and Yu Gu and Muhao Chen and Hoifung Poon},
booktitle={The Twelfth International Conference on Learning Representations},
year={2024},
url={https://openreview.net/forum?id=r65xfUb76p}
}

@inproceedings{lu-etal-2022-unified,
    title = "Unified Structure Generation for Universal Information Extraction",
    author = "Lu, Yaojie  and
      Liu, Qing  and
      Dai, Dai  and
      Xiao, Xinyan  and
      Lin, Hongyu  and
      Han, Xianpei  and
      Sun, Le  and
      Wu, Hua",
    editor = "Muresan, Smaranda  and
      Nakov, Preslav  and
      Villavicencio, Aline",
    booktitle = "Proceedings of the 60th Annual Meeting of the Association for Computational Linguistics (Volume 1: Long Papers)",
    month = may,
    year = "2022",
    address = "Dublin, Ireland",
    publisher = "Association for Computational Linguistics",
    url = "https://aclanthology.org/2022.acl-long.395/",
    doi = "10.18653/v1/2022.acl-long.395",
    pages = "5755--5772"
}

@inproceedings{NEURIPS2020_1457c0d6,
 author = {Brown, Tom and Mann, Benjamin and Ryder, Nick and Subbiah, Melanie and Kaplan, Jared D and Dhariwal, Prafulla and Neelakantan, Arvind and Shyam, Pranav and Sastry, Girish and Askell, Amanda and Agarwal, Sandhini and Herbert-Voss, Ariel and Krueger, Gretchen and Henighan, Tom and Child, Rewon and Ramesh, Aditya and Ziegler, Daniel and Wu, Jeffrey and Winter, Clemens and Hesse, Chris and Chen, Mark and Sigler, Eric and Litwin, Mateusz and Gray, Scott and Chess, Benjamin and Clark, Jack and Berner, Christopher and McCandlish, Sam and Radford, Alec and Sutskever, Ilya and Amodei, Dario},
 booktitle = {Advances in Neural Information Processing Systems},
 editor = {H. Larochelle and M. Ranzato and R. Hadsell and M.F. Balcan and H. Lin},
 pages = {1877--1901},
 publisher = {Curran Associates, Inc.},
 title = {Language Models are Few-Shot Learners},
 url = {https://proceedings.neurips.cc/paper_files/paper/2020/file/1457c0d6bfcb4967418bfb8ac142f64a-Paper.pdf},
 volume = {33},
 year = {2020}
}

@article{achiam2023gpt,
  title={{GPT}-4 technical report},
  author={OpenAI},
  journal={arXiv preprint arXiv:2303.08774},
url="https://arxiv.org/abs/2303.08774",
  year={2023}
}

@misc{openaichatgpt,
 author = {OpenAI},
 title = {{GPT}-3.5 Turbo Model},
url="https://platform.openai.com/docs/models/gpt-3-5-turbo",
 year = {2022}
}

@article{liu2024deepseek,
  title={Deepseek-{V}3 technical report},
  author={DeepSeek-AI},
  journal={arXiv preprint arXiv:2412.19437},
  url = "https://arxiv.org/abs/2412.19437",
  year={2024}
}

@inproceedings{balloccu-etal-2024-leak,
    title = "Leak, Cheat, Repeat: Data Contamination and Evaluation Malpractices in Closed-Source {LLM}s",
    author = "Balloccu, Simone  and
      Schmidtov{\'a}, Patr{\'i}cia  and
      Lango, Mateusz  and
      Dusek, Ondrej",
    editor = "Graham, Yvette  and
      Purver, Matthew",
    booktitle = "Proceedings of the 18th Conference of the European Chapter of the Association for Computational Linguistics (Volume 1: Long Papers)",
    month = mar,
    year = "2024",
    address = "St. Julian{'}s, Malta",
    publisher = "Association for Computational Linguistics",
    url = "https://aclanthology.org/2024.eacl-long.5/",
    doi = "10.18653/v1/2024.eacl-long.5",
    pages = "67--93"
}

@inproceedings{sainz-etal-2023-nlp,
    title = "{NLP} Evaluation in trouble: On the Need to Measure {LLM} Data Contamination for each Benchmark",
    author = "Sainz, Oscar  and
      Campos, Jon  and
      Garc{\'i}a-Ferrero, Iker  and
      Etxaniz, Julen  and
      de Lacalle, Oier Lopez  and
      Agirre, Eneko",
    editor = "Bouamor, Houda  and
      Pino, Juan  and
      Bali, Kalika",
    booktitle = "Findings of the Association for Computational Linguistics: EMNLP 2023",
    month = dec,
    year = "2023",
    address = "Singapore",
    publisher = "Association for Computational Linguistics",
    url = "https://aclanthology.org/2023.findings-emnlp.722/",
    doi = "10.18653/v1/2023.findings-emnlp.722",
    pages = "10776--10787"
}

@inproceedings{dong-etal-2024-generalization,
    title = "Generalization or Memorization: Data Contamination and Trustworthy Evaluation for Large Language Models",
    author = "Dong, Yihong  and
      Jiang, Xue  and
      Liu, Huanyu  and
      Jin, Zhi  and
      Gu, Bin  and
      Yang, Mengfei  and
      Li, Ge",
    editor = "Ku, Lun-Wei  and
      Martins, Andre  and
      Srikumar, Vivek",
    booktitle = "Findings of the Association for Computational Linguistics: ACL 2024",
    month = aug,
    year = "2024",
    address = "Bangkok, Thailand",
    publisher = "Association for Computational Linguistics",
    url = "https://aclanthology.org/2024.findings-acl.716/",
    doi = "10.18653/v1/2024.findings-acl.716",
    pages = "12039--12050"
}

@article{radford2019language,
  title={Language models are unsupervised multitask learners},
  author={Radford, Alec and Wu, Jeffrey and Child, Rewon and Luan, David and Amodei, Dario and Sutskever, Ilya and others},
  journal={OpenAI blog},
  volume={1},
  number={8},
  pages={9},
url="https://cdn.openai.com/better-language-models/language_models_are_unsupervised_multitask_learners.pdf",
  year={2019}
}

@article{walker2006ace,
  title={{ACE} 2005 multilingual training corpus},
  author={Walker, Christopher and Strassel, Stephanie and Medero, Julie and Maeda, Kazuaki},
  journal={Linguistic Data Consortium, Philadelphia},
  year={2006},
url="https://catalog.ldc.upenn.edu/LDC2006T06",
  publisher={Linguistic Data Consortium}
}

@inproceedings{li-etal-2020-dice,
    title = "Dice Loss for Data-imbalanced {NLP} Tasks",
    author = "Li, Xiaoya  and
      Sun, Xiaofei  and
      Meng, Yuxian  and
      Liang, Junjun  and
      Wu, Fei  and
      Li, Jiwei",
    editor = "Jurafsky, Dan  and
      Chai, Joyce  and
      Schluter, Natalie  and
      Tetreault, Joel",
    booktitle = "Proceedings of the 58th Annual Meeting of the Association for Computational Linguistics",
    month = jul,
    year = "2020",
    address = "Online",
    publisher = "Association for Computational Linguistics",
    url = "https://aclanthology.org/2020.acl-main.45/",
    doi = "10.18653/v1/2020.acl-main.45",
    pages = "465--476"
}

@article{cobbe2021trainingverifierssolvemath,
  title={Training verifiers to solve math word problems},
  author={Cobbe, Karl and Kosaraju, Vineet and Bavarian, Mohammad and Chen, Mark and Jun, Heewoo and Kaiser, Lukasz and Plappert, Matthias and Tworek, Jerry and Hilton, Jacob and Nakano, Reiichiro and others},
  journal={arXiv preprint arXiv:2110.14168},
  url={https://arxiv.org/abs/2110.14168}, 
  year={2021}
}

@inproceedings{zellers-etal-2019-hellaswag,
    title = "{H}ella{S}wag: Can a Machine Really Finish Your Sentence?",
    author = "Zellers, Rowan  and
      Holtzman, Ari  and
      Bisk, Yonatan  and
      Farhadi, Ali  and
      Choi, Yejin",
    editor = "Korhonen, Anna  and
      Traum, David  and
      M{\`a}rquez, Llu{\'i}s",
    booktitle = "Proceedings of the 57th Annual Meeting of the Association for Computational Linguistics",
    month = jul,
    year = "2019",
    address = "Florence, Italy",
    publisher = "Association for Computational Linguistics",
    url = "https://aclanthology.org/P19-1472/",
    doi = "10.18653/v1/P19-1472",
    pages = "4791--4800"
}

@inproceedings{dua-etal-2019-drop,
    title = "{DROP}: A Reading Comprehension Benchmark Requiring Discrete Reasoning Over Paragraphs",
    author = "Dua, Dheeru  and
      Wang, Yizhong  and
      Dasigi, Pradeep  and
      Stanovsky, Gabriel  and
      Singh, Sameer  and
      Gardner, Matt",
    editor = "Burstein, Jill  and
      Doran, Christy  and
      Solorio, Thamar",
    booktitle = "Proceedings of the 2019 Conference of the North {A}merican Chapter of the Association for Computational Linguistics: Human Language Technologies, Volume 1 (Long and Short Papers)",
    month = jun,
    year = "2019",
    address = "Minneapolis, Minnesota",
    publisher = "Association for Computational Linguistics",
    url = "https://aclanthology.org/N19-1246/",
    doi = "10.18653/v1/N19-1246",
    pages = "2368--2378"
}

@inproceedings{lin-etal-2022-truthfulqa,
    title = "{T}ruthful{QA}: Measuring How Models Mimic Human Falsehoods",
    author = "Lin, Stephanie  and
      Hilton, Jacob  and
      Evans, Owain",
    editor = "Muresan, Smaranda  and
      Nakov, Preslav  and
      Villavicencio, Aline",
    booktitle = "Proceedings of the 60th Annual Meeting of the Association for Computational Linguistics (Volume 1: Long Papers)",
    month = may,
    year = "2022",
    address = "Dublin, Ireland",
    publisher = "Association for Computational Linguistics",
    url = "https://aclanthology.org/2022.acl-long.229/",
    doi = "10.18653/v1/2022.acl-long.229",
    pages = "3214--3252"
}

@inproceedings{
hendrycks2021measuring,
title={Measuring Massive Multitask Language Understanding},
author={Dan Hendrycks and Collin Burns and Steven Basart and Andy Zou and Mantas Mazeika and Dawn Song and Jacob Steinhardt},
booktitle={International Conference on Learning Representations},
year={2021},
url={https://openreview.net/forum?id=d7KBjmI3GmQ}
}

@misc{eval-harness,
  author       = {Gao, Leo and Tow, Jonathan and Abbasi, Baber and Biderman, Stella and Black, Sid and DiPofi, Anthony and Foster, Charles and Golding, Laurence and Hsu, Jeffrey and Le Noac'h, Alain and Li, Haonan and McDonell, Kyle and Muennighoff, Niklas and Ociepa, Chris and Phang, Jason and Reynolds, Laria and Schoelkopf, Hailey and Skowron, Aviya and Sutawika, Lintang and Tang, Eric and Thite, Anish and Wang, Ben and Wang, Kevin and Zou, Andy},
  title        = {The Language Model Evaluation Harness},
  month        = 07,
  year         = 2024,
  publisher    = {Zenodo},
  version      = {v0.4.3},
  doi          = {10.5281/zenodo.12608602},
  url          = {https://zenodo.org/records/12608602}
}

@article{lv2025unified,
  title={A Unified Biomedical Named Entity Recognition Framework with Large Language Models},
  author={Lv, Tengxiao and Luo, Ling and Li, Juntao and Wang, Yanhua and Pan, Yuchen and Liu, Chao and Wang, Yanan and Jiang, Yan and Lv, Huiyi and Sun, Yuanyuan and others},
  journal={arXiv preprint arXiv:2510.08902},
  url={https://arxiv.org/abs/2510.08902},
  year={2025}
}

@article{gemmateam2025gemma3technicalreport,
  title={Gemma 3 Technical Report}, 
  author={{Gemma Team}},
  journal={arXiv preprint arXiv:2503.19786},
  url={https://arxiv.org/abs/2503.19786},
  year={2025}, 
}

@misc{granite2025,
  author       = {{IBM Research}},
  title        = {Granite 4.0 Language Models},
  year         = {2025},
  howpublished = {\url{https://github.com/ibm-granite/granite-4.0-language-models}},
  note         = {Accessed: 2025-10-01}
}

@inproceedings{rucker-akbik-2023-cleanconll,
    title = "{C}lean{C}o{NLL}: A Nearly Noise-Free Named Entity Recognition Dataset",
    author = {R{\"u}cker, Susanna  and
      Akbik, Alan},
    editor = "Bouamor, Houda  and
      Pino, Juan  and
      Bali, Kalika",
    booktitle = "Proceedings of the 2023 Conference on Empirical Methods in Natural Language Processing",
    month = dec,
    year = "2023",
    address = "Singapore",
    publisher = "Association for Computational Linguistics",
    url = "https://aclanthology.org/2023.emnlp-main.533/",
    doi = "10.18653/v1/2023.emnlp-main.533",
    pages = "8628--8645"
}

\appendix

\section{Checkpoints of Models}
\label{app:model_checkpoints}
Table~\ref{table:model_checkpoints} lists the checkpoints of the open-source large language models used in our experiments.

\begin{table}[H]
    \centering
    \small
    \renewcommand{\arraystretch}{1.15}
    \begin{tabularx}{\columnwidth}{@{}>{\raggedright\arraybackslash}p{0.28\columnwidth}>{\raggedright\arraybackslash}X@{}}
        \toprule
        \textbf{Model} & \textbf{Resource Link} \\
        \midrule
        \addlinespace[1pt]
        \rowcolor{gray!20}
        \multicolumn{2}{c}{\llamaLogo\textit{LLaMA Family}} \\
        LLaMA3.1-8B & \href{https://huggingface.co/meta-llama/Llama-3.1-8B-Instruct}{meta-llama/Llama-3.1-8B-Instruct} \\
        LLaMA3.2-3B & \href{https://huggingface.co/meta-llama/Llama-3.2-3B-Instruct}{meta-llama/Llama-3.2-3B-Instruct} \\
        \midrule
        \addlinespace[1pt]
        \rowcolor{gray!20}
        \multicolumn{2}{c}{\qwenLogo\textit{Qwen Family}} \\
        Qwen2.5-7B & \href{https://huggingface.co/Qwen/Qwen2.5-7B-Instruct}{Qwen/Qwen2.5-7B-Instruct} \\
        Qwen3-4B & \href{https://huggingface.co/Qwen/Qwen3-4B-Instruct-2507}{Qwen/Qwen3-4B-Instruct-2507} \\
        \midrule
        \addlinespace[1pt]
        \rowcolor{gray!20}
        \multicolumn{2}{c}{\graniteLogo\textit{Granite Family}} \\
        Granite3.3-8B & \href{https://huggingface.co/ibm-granite/granite-3.3-8b-instruct}{ibm-granite/granite-3.3-8b-instruct} \\
        Granite4-3B & \href{https://huggingface.co/ibm-granite/granite-4.0-micro}{ibm-granite/granite-4.0-micro} \\
        \midrule
        \addlinespace[1pt]
        \rowcolor{gray!20}
        \multicolumn{2}{c}{\geminiLogo\textit{Gemma Family}} \\
        Gemma2-9B & \href{https://huggingface.co/google/gemma-2-9b-it}{google/gemma-2-9b-it} \\
        Gemma3-4B & \href{https://huggingface.co/google/gemma-3-4b-it}{google/gemma-3-4b-it} \\
        \bottomrule
    \end{tabularx}
    \caption{Checkpoints of open-source models used in our experiments.}
    \label{table:model_checkpoints}
\end{table}

\section{Full Instructions for Different Datasets}
\label{app:prompt}
We provide the complete instruction design used for LLMs on generative NER tasks below.

\begin{figure}[t!]
    \centering
    \begin{tcolorbox}[
        colback=white,
        colframe=black,
        colbacktitle=black,
        center title,
        fonttitle=\bfseries,
        top=10pt,
        title=Instruction Framework,
    ]
        
        {
        \textbf{[Task Description]}\\ 
        Your task is to identify all named entities in the input sentence and rewrite the sentence by enclosing each entity using the format <LABEL>Entity Text</LABEL>. Use only the label tags defined in the Label Set below.\\
        }

        {
        \textbf{[Label Definition Schema]}\\
        Label Set:\\
        \textbf{\textsc{ORG}(organization):} A collective entity such as a company, institution, brand, political or governmental body, publication, or any organized group of people acting as a unit.\\
        \textbf{\textsc{PER}(person):} A named individual, including humans, animals, fictional characters, and their aliases.\\
        \textbf{\textsc{LOC}(location):} A geographical or spatial entity, including natural features, built structures, regions, public or commercial places, assorted buildings, and abstract or metaphorical places.\\
        \textbf{\textsc{MISC}(miscellaneous):} Named entities that are not persons, organizations, or locations, including derived adjectives, religions, ideologies, nationalities, languages, events, programs, wars, titles of works, slogans, eras, and types of objects.\\
        }
        
        {
        \textbf{[Input Sentence]}\\
        Now process the input sentence:
        }
        
    \end{tcolorbox}
    \caption{The template and example of constructing instructions for generative NER.}
    \label{fig:instruction_framework}
\end{figure}

\subsection{Instruction Construction Framework}
Each instruction follows a unified structural organization. As illustrated in Figure~\ref{fig:instruction_framework}, the constructed input is composed of three main components: \textbf{Task Description}, which defines the extraction goal and output format; \textbf{Label Definition Schema}, which provides fine-grained semantic constraints for each entity type; \textbf{Input Sentence}, the target text for labeling. These components collectively guide the model to generate the target output according to the specified format.

\subsection{Task Descriptions for Different Output Formats}
We designed five distinct task descriptions corresponding to the output formats. These descriptions explicitly constrain the generation behavior of the LLMs. Table~\ref{tab:task_descriptions} details the specific prompt templates used for each format.

\begin{table*}[t]
\centering
\renewcommand{\arraystretch}{1.2}
\begin{tabularx}{\textwidth}{lX}
\toprule
\textbf{Output Format} & \textbf{Task Description} \\
\midrule

Inline Bracketed &
Your task is to identify all named entities in the input sentence and rewrite the sentence by enclosing each entity using the format [Entity Text | LABEL]. Use only the label tags defined in the Label Set below. \\

\midrule

Inline XML &
Your task is to identify all named entities in the input sentence and rewrite the sentence by enclosing each entity using the format <LABEL>Entity Text</LABEL>. Use only the label tags defined in the Label Set below. \\
\midrule

Category-grouped JSON &
Your task is to identify all named entities in the input sentence and extract them into a JSON object grouped by label. The keys of the JSON object should be the label tags defined in the Label Set below, and the values should be lists of extracted entities. If a label has no corresponding entities, use an empty list. \\

\midrule

Occurrence-based JSON &
Your task is to identify all named entities in the input sentence and extract them into a JSON list. Output a JSON list where each object contains:

1. ``text'': the entity string.

2. ``label'': the label tags defined in the Label Set below.

3. ``occurrence\_index'': a 1-based integer indicating the order of appearance of this text in the sentence (e.g., 1 for the first appearance, 2 for the second).

If no entities are found, return an empty list.  \\

\midrule

Offset-based JSON &
Your task is to identify all named entities in the input sentence and extract them with precise character offsets. Output a JSON list where each object contains:

1. ``text'': the entity string.

2. ``label'': the label tags defined in the Label Set below.

3. ``start'': the 0-based index of the entity's first character.

4. ``end'': the 0-based index of the first character after the entity (exclusive). \\

\bottomrule
\end{tabularx}

\caption{Task descriptions for different output formats.}
\label{tab:task_descriptions}
\end{table*}

\subsection{Label Definition Schema}
\label{app:label_definition}
To enhance the semantic understanding of the LLM, particularly for ambiguous or domain-specific entities, we provide explicit definitions for each label in the instructions. These definitions are derived from the official annotation guidelines of each dataset, which serve as the authoritative standard in the dataset's labeling process. For the CoNLL2003 dataset, the official annotation guidelines contain extensive explanations for the labels, which we have summarized using a large language model to provide concise, standard linguistic definitions. For the GENIA dataset, the official annotation guidelines were not available; therefore, we employed a large language model to generate standard biomedical definitions. The detailed definitions for CoNLL2003, OntoNotes5.0, GENIA, and ACE2005 are provided in Table~\ref{tab:conll_def}, Table~\ref{tab:onto_def}, Table~\ref{tab:genia_def}, and Table~\ref{tab:ace_def}, respectively.

\section{Brief Introduction of Baselines}
\label{app:baselines}
\textbf{BERT-Tagger}~\cite{devlin-etal-2019-bert} uses a BERT model and trains classification heads for tagging tokens. \textbf{GNN-SL}~\cite{wang-etal-2023-gnn} augments sequence labeling by retrieving nearest neighbor examples and modeling their interactions with the input via GNNs. \textbf{ACE + Fine-tune}~\cite{wang-etal-2021-automated} utilizes a reinforcement learning controller to automatically search for the optimal concatenation of various pre-trained embeddings. For a fair comparison, we adopt its sentence-level variant, achieving SOTA performance on the CoNLL2003 dataset. \textbf{BERT-MRC+DSC}~\cite{li-etal-2020-dice} addresses data imbalance via Dice loss, achieving SOTA performance on the OntoNotes5.0 dataset. \textbf{BERT-MRC}~\cite{li-etal-2020-unified} formulates NER as a machine reading comprehension task, achieving SOTA performance on the GENIA dataset. \textbf{BINDER}~\cite{zhang2023optimizing} aligns spans and types in a shared space via a bi-encoder and contrastive learning, distinguishing entities with dynamic thresholding. \textbf{W2NER}~\cite{Li_Fei_Liu_Wu_Zhang_Teng_Ji_Li_2022} unifies flat, nested, and discontinuous NER by modeling word-word relations on a 2D grid to capture token interactions. \textbf{Biaffine+CNN}~\cite{yan-etal-2023-embarrassingly} treats the span feature grid generated by a biaffine decoder as an image, utilizing CNNs to model spatial correlations between neighboring spans for nested entity recognition. 

\section{Comparison between LoRA and Full Fine-Tuning}
\label{app:lora_full}

We compare LoRA with full fine-tuning to determine which strategy is more effective for generative NER. The two well-performing models, LLaMA3.1-8B and Qwen2.5-7B, are fine-tuned on the flat NER dataset (CoNLL2003) and the nested NER dataset (GENIA). Both fine-tuning methods use the same training configuration. The F1 results are shown in Table~\ref{table:lora_full}.

\begin{table}[t]
    \centering
    \small
    \resizebox{\columnwidth}{!}{
    \begin{tabular}{llcc}
    \toprule
    \textbf{Model} & \textbf{Fine-tuning} & \textbf{CoNLL2003} & \textbf{GENIA} \\
    \midrule
    LLaMA3.1-8B & LoRA & 93.85 & 79.18 \\
    LLaMA3.1-8B & Full & 93.61~\textcolor{magenta}{($\downarrow$0.24)} & 78.70~\textcolor{magenta}{($\downarrow$0.48)} \\
    Qwen2.5-7B  & LoRA & 93.73 & 78.67 \\
    Qwen2.5-7B  & Full & 93.58~\textcolor{magenta}{($\downarrow$0.15)} & 78.53~\textcolor{magenta}{($\downarrow$0.14)} \\
    \bottomrule
    \end{tabular}}
    \caption{Comparison between LoRA fine-tuning and full fine-tuning on CoNLL2003 and GENIA. Numbers in parentheses denote the difference compared with LoRA fine-tuning.}
    \label{table:lora_full}
    \end{table}

    Across all settings, LoRA fine-tuning obtains slightly higher F1 scores than full fine-tuning. Moreover, because the number of trainable parameters in LoRA is much closer to that of traditional NER models, LoRA also provides a fairer basis for comparing generative and traditional approaches. Therefore, we adopt LoRA as the fine-tuning strategy for all LLMs in our main experiments.

\section{Breakdown of Flat and Nested Entity Performance}
\label{app:flat_nested_breakdown}

We further break down performance on the two nested NER datasets, ACE2005 and GENIA, according to the structural relation between entity spans. We define a \textit{nested entity} as a gold entity whose span strictly contains or is strictly contained by another gold entity in the same sentence. A \textit{flat entity} neither contains nor is contained by another gold entity. We compute entity-level Micro-F1 separately for these two disjoint subsets and average the scores across all eight LLMs.

\begin{figure}[t!]
    \centering
    \includegraphics[width=\columnwidth]{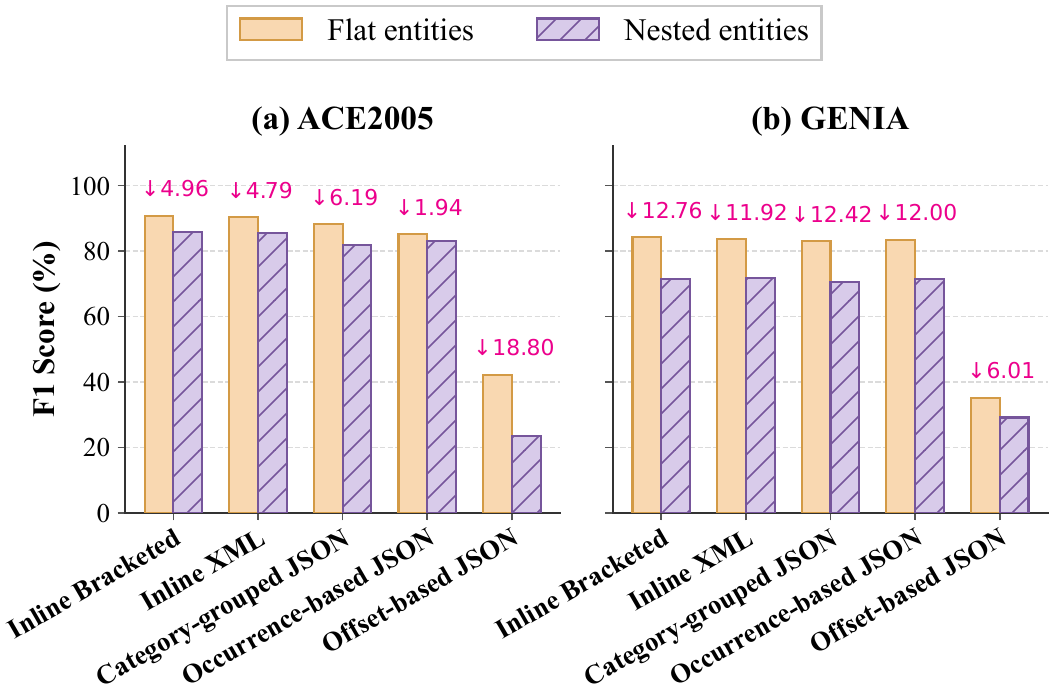}
    \caption{Performance on flat and nested entity mentions in the two nested NER datasets, averaged across eight LLMs for each output format. Downward arrows report the F1 decrease from flat to nested entities.}
    \label{fig:flat_nested_breakdown}
\end{figure}

As shown in Figure~\ref{fig:flat_nested_breakdown}, nested entities consistently achieve lower F1 scores than flat entities across all output formats and on both datasets, indicating that they are more difficult to recognize. Excluding the Offset-based JSON format, whose overall 
scores are too low for meaningful comparison, the 
flat-to-nested F1 decrease on GENIA is consistently 
larger than on ACE2005, suggesting that nested entity 
recognition is particularly challenging in the 
biomedical domain.

\begin{table*}[t!]
    \centering
    \small
    \setlength{\tabcolsep}{5pt}
    \renewcommand{\arraystretch}{1.15}
    \resizebox{\textwidth}{!}{%
    \begin{tabular}{@{}l cc cc cc cc@{}}
        \toprule
        \multirow{2}{*}{\textbf{Model}} & \multicolumn{2}{c}{\textbf{CoNLL2003}} & \multicolumn{2}{c}{\textbf{OntoNotes5.0}} & \multicolumn{2}{c}{\textbf{ACE2005}} & \multicolumn{2}{c}{\textbf{GENIA}} \\
        \cmidrule(lr){2-3} \cmidrule(lr){4-5} \cmidrule(lr){6-7} \cmidrule(lr){8-9}
        & F1 & $F_1^{\text{w/o offset}}$ & F1 & $F_1^{\text{w/o offset}}$ & F1 & $F_1^{\text{w/o offset}}$ & F1 & $F_1^{\text{w/o offset}}$ \\
        \midrule
        LLaMA3.1-8B & 63.22 & 92.64~\textcolor{teal}{($\uparrow$29.42)} & 53.31 & 87.60~\textcolor{teal}{($\uparrow$34.29)} & 35.85 & 81.60~\textcolor{teal}{($\uparrow$45.75)} & 34.62 & 77.96~\textcolor{teal}{($\uparrow$43.34)} \\
        LLaMA3.2-3B & 56.13 & 91.91~\textcolor{teal}{($\uparrow$35.78)} & 31.20 & 84.71~\textcolor{teal}{($\uparrow$53.51)} & 27.31 & 79.54~\textcolor{teal}{($\uparrow$52.23)} & 25.79 & 76.63~\textcolor{teal}{($\uparrow$50.84)} \\
        Qwen2.5-7B & 60.84 & 92.74~\textcolor{teal}{($\uparrow$31.90)} & 50.92 & 87.59~\textcolor{teal}{($\uparrow$36.67)} & 32.59 & 80.61~\textcolor{teal}{($\uparrow$48.02)} & 32.02 & 76.45~\textcolor{teal}{($\uparrow$44.43)} \\
        Qwen3-4B & 61.18 & 91.97~\textcolor{teal}{($\uparrow$30.79)} & 38.34 & 84.76~\textcolor{teal}{($\uparrow$46.42)} & 35.41 & 80.07~\textcolor{teal}{($\uparrow$44.66)} & 36.81 & 77.33~\textcolor{teal}{($\uparrow$40.52)} \\
        Granite3.3-8B & 64.79 & 92.90~\textcolor{teal}{($\uparrow$28.11)} & 44.08 & 87.05~\textcolor{teal}{($\uparrow$42.97)} & 36.90 & 81.86~\textcolor{teal}{($\uparrow$44.96)} & 39.54 & 77.50~\textcolor{teal}{($\uparrow$37.96)} \\
        Granite4-3B & 61.40 & 92.40~\textcolor{teal}{($\uparrow$31.00)} & 38.91 & 86.85~\textcolor{teal}{($\uparrow$47.94)} & 33.30 & 80.95~\textcolor{teal}{($\uparrow$47.65)} & 36.80 & 77.01~\textcolor{teal}{($\uparrow$40.21)} \\
        Gemma2-9B & 65.53 & 92.47~\textcolor{teal}{($\uparrow$26.94)} & 61.45 & 89.35~\textcolor{teal}{($\uparrow$27.90)} & 40.80 & 83.30~\textcolor{teal}{($\uparrow$42.50)} & 41.15 & 77.93~\textcolor{teal}{($\uparrow$36.78)} \\
        Gemma3-4B & 59.85 & 92.24~\textcolor{teal}{($\uparrow$32.39)} & 51.35 & 86.28~\textcolor{teal}{($\uparrow$34.93)} & 31.60 & 79.11~\textcolor{teal}{($\uparrow$47.51)} & 32.00 & 75.65~\textcolor{teal}{($\uparrow$43.65)} \\
        \hdashline
        \rowcolor{myrowblue}
        \textbf{Average} & 61.62 & 92.41~\textcolor{teal}{($\uparrow$30.79)} & 46.20 & 86.77~\textcolor{teal}{($\uparrow$40.58)} & 34.22 & 80.88~\textcolor{teal}{($\uparrow$46.66)} & 34.84 & 77.06~\textcolor{teal}{($\uparrow$42.22)} \\
        \bottomrule
    \end{tabular}}
    \caption{Impact of character-offset errors on Offset-based JSON 
    performance. $F_1^{\text{w/o offset}}$ removes the offset-matching 
    requirement, revealing that models identify entities accurately 
    but struggle with precise positional alignment. Green values indicate the absolute F1 improvement 
    from ignoring offsets. }
    \label{table:offset_recognition}
\end{table*}

\section{Error Analysis of the Offset-based JSON Format}
\label{app:offset_error_analysis}

The Offset-based JSON format yields substantially lower F1 scores than the other formats, as reported in Tables~\ref{table:flat_ner_results} and~\ref{table:nested_ner_results}. To further investigate the reasons for the low Offset-based JSON performance, we present two complementary analyses below.

\paragraph{Entity recognition vs.\ offset generation.}
We re-evaluate predictions under two criteria. Standard F1 requires an exact match of the entity text, label, and character offsets. $F_1^{\text{w/o offset}}$ matches only the entity text and label. As shown in Table~\ref{table:offset_recognition}, ignoring offsets leads to large improvements across all models and datasets, with performance approaching that of the inline formats. This confirms that the models usually accurately identify entities, while the primary difficulty lies in precise character-level offset alignment.

\begin{figure}[!ht]
    \centering
    \includegraphics[width=\columnwidth]{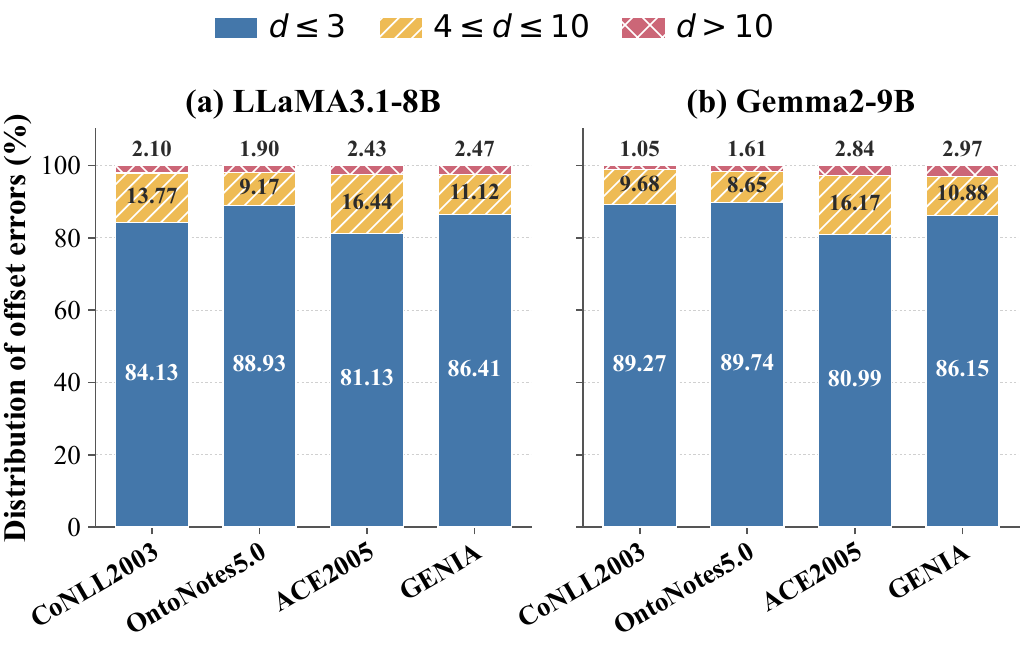}
    \caption{Distribution of character-offset errors by boundary displacement $d$ for two representative models across four datasets.}
    \label{fig:offset_displacement}
\end{figure}

\paragraph{Magnitude of character-offset errors.}
We further examine predictions whose entity text and label are correct but whose offsets are incorrect. For each prediction, we define the boundary displacement as $d=\max(|\hat{s}-s|,|\hat{e}-e|)$, where $(\hat{s},\hat{e})$ and $(s,e)$ are the predicted and gold start and end offsets, respectively. As shown in Figure~\ref{fig:offset_displacement}, across all models and datasets, 81–90\% of offset errors fall within 3 characters, and large deviations (>10 characters) account for only 1–3\%.

Together, these analyses confirm that \textbf{the low Offset-based JSON scores are caused by character-level offset misalignment, not by failures in entity recognition}, and that most errors involve only small boundary deviations. These results indicate that exact character indexing is difficult for autoregressive language models. Clearer instructions and additional examples may reduce some errors, but they are unlikely to remove the problem entirely. 

\section{Cost-Benefit Analysis}
\label{app:cost-benefit_analysis}
    
We further conduct a cost-benefit experiment between the best-performing generative NER model, LLaMA3.1-8B with the Inline Bracketed format, and a representative traditional NER model, BERT-MRC, on CoNLL2003. Both models were trained on a single NVIDIA L40 GPU (48GB). The results are shown in Table~\ref{table:cost_benefit}.

\begin{table}[t]
    \centering
    \footnotesize
    \setlength{\tabcolsep}{4pt}
    \renewcommand{\arraystretch}{1.12}
    \begin{tabular}{@{}lrr@{}}
    \toprule
    \textbf{Metric}
        & \multicolumn{1}{c}{\textbf{LLaMA3.1-8B}}
        & \multicolumn{1}{c}{\textbf{BERT-MRC}} \\
    \midrule
    Training Time (h)
        & 2.40 & 2.31 \\
    Trainable Parameters (M)
        & 671 & 335 \\
    Peak GPU Memory (GiB)
        & 27.89 & 21.62 \\
    \addlinespace[2pt]
    F1
        & 93.85 & 93.04 \\
    \bottomrule
    \end{tabular}
    \caption{Cost-benefit comparison between LLaMA3.1-8B and BERT-MRC on CoNLL2003.}
    \label{table:cost_benefit}
\end{table}
    
While LLaMA3.1-8B requires slightly more resources than BERT-MRC, the gap is modest considering its strong performance (93.85 vs. 93.04). This trade-off demonstrates the viability of generative NER under resource-efficient fine-tuning.

\section{Full Precision, Recall, and F1 Results}
\label{app:full_prf_results}

To provide a complete view of model performance, we report Precision, Recall, and F1 scores for the flat and nested NER experiments in Table~\ref{table:full_prf_results}.

\section{Error Type Definitions}
\label{app:error_definitions}

Prediction errors are categorized into eight distinct types. The detailed definitions are as follows:

\noindent\textbf{Type Errors.} These errors involve correct boundaries but incorrect labeling. \textit{OOD Types} refer to predicted entity types that do not exist in the provided label set. \textit{Wrong Types} denote cases where the predicted type is incorrect but belongs to the given label set.

\noindent\textbf{Boundary Errors.} These occur when the predicted span is imprecise. \textit{Contain Gold} describes predicted mentions that enclose the gold mentions. \textit{Contained by Gold} refers to predicted mentions that are inside the gold mentions. \textit{Overlap with Gold} covers predictions that overlap with gold mentions but do not fall into the above two cases.

\noindent\textbf{Other Errors.} \textit{Completely-O} represents predicted mentions that have no overlap with any gold mentions. \textit{OOD Mentions} correspond to hallucinated predictions that do not appear in the input text. \textit{Omitted Mentions} refer to gold entities that the model fails to identify.

\section{AI Assistance Statement}
\label{app:ai_statement}

We utilized LLMs (e.g., ChatGPT) to assist in polishing the text, refining grammar, and improving the clarity of the manuscript. All content was reviewed and verified by the authors.

As detailed in Appendix~\ref{app:label_definition}, we employed LLMs to summarize official annotation guidelines for CoNLL2003 dataset and generate standard biomedical definitions for GENIA dataset to construct the label definition schema used in our instruction tuning prompts.

\begin{table*}[t]
\centering
\renewcommand{\arraystretch}{1.3}
\begin{tabularx}{\textwidth}{l X}
\toprule
\textbf{Labels} & \textbf{Definition} \\
\midrule
\textsc{ORG} & A collective entity such as a company, institution, brand, political or governmental body, publication, or any organized group of people acting as a unit. \\
\textsc{PER} & A named individual, including humans, animals, fictional characters, and their aliases. \\
\textsc{LOC} & A geographical or spatial entity, including natural features, built structures, regions, public or commercial places, assorted buildings, and abstract or metaphorical places. \\
\textsc{MISC} & Named entities that are not persons, organizations, or locations, including derived adjectives, religions, ideologies, nationalities, languages, events, programs, wars, titles of works, slogans, eras, and types of objects. \\
\bottomrule
\end{tabularx}
\caption{Label definitions for CoNLL2003 dataset.}
\label{tab:conll_def}
\end{table*}

\begin{table*}[ht]
\centering
\renewcommand{\arraystretch}{1.3}
\begin{tabular}{ l p{9cm} }
\toprule
\textbf{Labels} & \textbf{Definition} \\
\midrule
\textsc{PERSON} & People, including fictional. \\
\textsc{NORP} & Nationalities or religious or political groups. \\
\textsc{FAC} & Buildings, airports, highways, bridges, etc.. \\
\textsc{ORG} & Companies, agencies, institutions, etc.. \\
\textsc{GPE} & Countries, cities, states. \\
\textsc{LOC} & Non-GPE locations, mountain ranges, bodies of water. \\
\textsc{PRODUCT} & Vehicles, weapons, foods, etc. (Not services). \\
\textsc{EVENT} & Named hurricanes, battles, wars, sports events, etc.. \\
\textsc{WORK\_OF\_ART} & Titles of books, songs, etc.. \\
\textsc{LAW} & Named documents made into laws. \\
\textsc{LANGUAGE} & Any named language. \\
\textsc{DATE} & Absolute or relative dates or periods. \\
\textsc{TIME} & Times smaller than a day. \\
\textsc{PERCENT} & Percentage (including ``\%''). \\
\textsc{MONEY} & Monetary values, including unit. \\
\textsc{QUANTITY} & Measurements, as of weight or distance. \\
\textsc{ORDINAL} & ``first'', ``second''. \\
\textsc{CARDINAL} & Numerals that do not fall under another type. \\
\bottomrule
\end{tabular}
\caption{Label definitions for OntoNotes5.0 dataset.}
\label{tab:onto_def}
\end{table*}

\begin{table*}[ht]
\centering
\renewcommand{\arraystretch}{1.3}
\begin{tabularx}{\textwidth}{l X}
\toprule
\textbf{Labels} & \textbf{Definition} \\
\midrule
\textsc{DNA} & Deoxyribonucleic acid sequences or molecules, including genes, gene fragments, promoters, enhancers, and other genomic regions that encode or regulate biological information. \\
\textsc{RNA} & Ribonucleic acid molecules transcribed from DNA, encompassing messenger RNA, non-coding RNAs, and RNA transcripts involved in gene expression and regulation. \\
\textsc{Protein} & Functional biological macromolecules composed of amino acids, including enzymes, receptors, transcription factors, and protein complexes produced by translation of RNA. \\
\textsc{Cell\_line} & A population of cells derived from a single source and maintained in vitro through continuous culture, often genetically stable and used for experimental research. \\
\textsc{Cell\_type} & A class of cells defined by shared morphological, functional, and molecular characteristics within an organism, such as immune cells, epithelial cells, or neurons. \\
\bottomrule
\end{tabularx}
\caption{Label definitions for GENIA dataset.}
\label{tab:genia_def}
\end{table*}

\begin{table*}[ht]
\centering
\renewcommand{\arraystretch}{1.3}
\begin{tabularx}{\textwidth}{l X}
\toprule
\textbf{Labels} & \textbf{Definition} \\
\midrule
\textsc{PER} & Person entities are limited to humans. A person may be a single individual or a group. \\
\textsc{ORG} & Organization entities are limited to corporations, agencies, and other groups of people defined by an established organizational structure. \\
\textsc{GPE} & GPE entities are geographical regions defined by political and/or social groups. A GPE entity subsumes and does not distinguish between a nation, its region, its government, or its people. \\
\textsc{LOC} & Location entities are limited to geographical entities such as geographical areas and landmasses, bodies of water, and geological formations. \\
\textsc{FAC} & Facility entities are limited to buildings and other permanent man-made structures and real estate improvements. \\
\textsc{VEH} & A vehicle entity is a physical device primarily designed to move an object from one location to another, by (for example) carrying, pulling, or pushing the transported object. Vehicle entities may or may not have their own power source.\\
\textsc{WEA} & Weapon entities are limited to physical devices primarily used as instruments for physically harming or destroying other entities. \\
\bottomrule
\end{tabularx}
\caption{Label definitions for ACE2005 dataset.}
\label{tab:ace_def}
\end{table*}

\begin{table*}[t!]
\small
	\centering
	\scalebox{0.82}{
    \renewcommand{\arraystretch}{1.15}
	\begin{tabular}{lcccccccccccccc}
		\toprule
		\multirow{2}{*}{\textbf{Models}} & \multicolumn{3}{c}{\textbf{CoNLL2003}} & \multicolumn{3}{c}{\textbf{OntoNotes5.0}} & \multirow{2}{*}{\textbf{Avg}$^{\text{Flat}}_{\text{F1}}$} & \multicolumn{3}{c}{\textbf{ACE2005}} & \multicolumn{3}{c}{\textbf{GENIA}} & \multirow{2}{*}{\textbf{Avg}$^{\text{Nested}}_{\text{F1}}$} \\
		\cmidrule(lr){2-4}\cmidrule(lr){5-7}\cmidrule(lr){9-11}\cmidrule(lr){12-14}
		& P & R & F1 & P & R & F1 & & P & R & F1 & P & R & F1 & \\
		\midrule
		\addlinespace[1pt]
		\rowcolor{gray!20}
		\multicolumn{15}{c}{\textit{Pre-Trained NER Models}} \\
		BERT-Tagger & 91.20 & 92.30 & 91.70 & 90.01 & 88.35 & 89.16 & 90.43 & - & - & - & - & - & - & - \\
		GNN-SL & 93.02 & 93.40 & 93.20 & 91.48 & 91.29 & 91.39 & 92.30 & - & - & - & - & - & - & - \\
		ACE + Fine-tune & 93.07 & 94.21 & 93.64 & 91.48 & 91.29 & 91.39 & 92.52 & - & - & - & - & - & - & - \\
        BERT-MRC+DSC & 93.41 & 93.25 & 93.33 & 91.59 & 92.56 & 92.07 & 92.70 & - & - & - & - & - & - & - \\
		BERT-MRC & 92.33 & 94.61 & 93.04 & 92.98 & 89.95 & 91.11 & 92.08 & 87.16 & 86.59 & 86.88 & 85.18 & 81.12 & 83.75 & 85.32 \\
		BINDER & - & - & - & - & - & - & - & 89.60 & 90.50 & 90.00 & 83.40 & 78.30 & 80.80 & 85.40 \\
		W2NER & - & - & - & - & - & - & - & 85.03 & 88.62 & 86.79 & 81.58 & 79.11 & 80.32 & 83.56 \\
		Biaffine+CNN & - & - & - & - & - & - & - & 86.78 & 87.72 & 87.25 & 81.52 & 79.17 & 80.33 & 83.79 \\
		\midrule
		\addlinespace[1pt]
		\rowcolor{gray!20}
		\multicolumn{15}{c}{\textit{GPT-NER}} \\
		GPT-3 (\textit{ZS}) & 88.54 & 91.40 & 89.97 & 79.17 & 84.29 & 81.73 & 85.85 & 71.72 & 74.20 & 72.96 & 61.38 & 66.74 & 64.06 & 68.51 \\
		GPT-3 (\textit{ZS w/ Ver.}) & 89.47 & 91.77 & 90.62 & 79.64 & 84.52 & 82.08 & 86.35 & 72.63 & 75.39 & 73.46 & 61.77 & 66.81 & 64.29 & 68.88 \\
		GPT-3 (\textit{FS w/ Ver.}) & 89.76 & 92.06 & 90.91 & 79.89 & 84.51 & 82.20 & 86.56 & 72.77 & 75.51 & 73.59 & 61.89 & 66.95 & 64.42 & 69.01 \\
		\midrule
		\addlinespace[1pt]
		\rowcolor{gray!20}
		\multicolumn{15}{c}{\textit{Large Language Models (Inline Bracketed Format)}} \\
		LLaMA3.1-8B & 93.50 & 94.21 & 93.85 & 91.23 & 91.32 & 91.28 & 92.57 & 87.16 & 87.85 & 87.51 & 81.93 & 76.62 & 79.18 & 83.35 \\
		LLaMA3.2-3B & 93.17 & 93.91 & 93.54 & 89.18 & 89.21 & 89.19 & 91.37 & 85.18 & 86.47 & 85.82 & 81.77 & 76.69 & 79.15 & 82.49 \\
		Qwen2.5-7B & 93.51 & 93.94 & 93.73 & 90.98 & 90.97 & 90.97 & 92.35 & 86.47 & 86.93 & 86.70 & 81.76 & 75.82 & 78.67 & 82.69 \\
		Qwen3-4B & 93.05 & 93.68 & 93.37 & 89.25 & 89.35 & 89.30 & 91.34 & 85.56 & 87.43 & 86.48 & 81.94 & 76.56 & 79.16 & 82.82 \\
		Granite3.3-8B & 93.38 & 94.14 & 93.76 & 90.37 & 90.26 & 90.31 & 92.04 & 86.08 & 87.56 & 86.81 & 81.24 & 76.64 & 78.87 & 82.84 \\
		Granite4-3B & 92.75 & 93.56 & 93.15 & 89.59 & 90.05 & 89.82 & 91.49 & 85.93 & 87.66 & 86.78 & 81.09 & 75.91 & 78.42 & 82.60 \\
		Gemma2-9B & 93.24 & 94.03 & 93.64 & 90.72 & 91.11 & 90.91 & 92.28 & 86.55 & 88.35 & 87.44 & 82.41 & 77.22 & 79.73 & 83.59 \\
		Gemma3-4B & 93.32 & 94.03 & 93.68 & 90.01 & 90.17 & 90.09 & 91.89 & 85.40 & 87.03 & 86.20 & 81.75 & 75.76 & 78.64 & 82.42 \\
		\hdashline
        \rowcolor{myrowblue}
		\textbf{Average} & 93.24 & 93.94 & 93.59 & 90.17 & 90.30 & 90.23 & \flatAvg{91.91} & 86.04 & 87.41 & 86.72 & 81.74 & 76.40 & 78.98 & \nestedAvg{82.85} \\
        
		\midrule
		\addlinespace[1pt]
		\rowcolor{gray!20}
		\multicolumn{15}{c}{\textit{Large Language Models (Inline XML Format)}} \\
		LLaMA3.1-8B & 93.57 & 94.09 & 93.83 & 90.84 & 91.05 & 90.94 & 92.39 & 85.30 & 86.98 & 86.13 & 80.78 & 76.18 & 78.41 & 82.27 \\
		LLaMA3.2-3B & 92.80 & 93.33 & 93.06 & 89.14 & 89.46 & 89.30 & 91.18 & 85.31 & 87.24 & 86.26 & 80.21 & 76.32 & 78.22 & 82.24 \\
		Qwen2.5-7B & 93.31 & 93.87 & 93.59 & 90.88 & 90.71 & 90.79 & 92.19 & 85.50 & 86.45 & 85.97 & 81.31 & 77.34 & 79.27 & 82.62 \\
		Qwen3-4B & 92.95 & 93.36 & 93.15 & 88.99 & 88.93 & 88.96 & 91.06 & 84.89 & 86.45 & 85.66 & 81.39 & 77.88 & 79.60 & 82.63 \\
		Granite3.3-8B & 92.91 & 93.75 & 93.33 & 89.87 & 90.04 & 89.95 & 91.64 & 86.87 & 87.67 & 87.27 & 81.42 & 76.89 & 79.09 & 83.18 \\
		Granite4-3B & 92.75 & 93.38 & 93.07 & 89.33 & 89.58 & 89.46 & 91.27 & 85.50 & 88.23 & 86.84 & 81.07 & 76.30 & 78.61 & 82.73 \\
		Gemma2-9B & 93.14 & 94.03 & 93.59 & 90.98 & 91.01 & 90.99 & 92.29 & 87.24 & 88.82 & 88.02 & 81.93 & 76.87 & 79.32 & 83.67 \\
		Gemma3-4B & 92.73 & 93.71 & 93.22 & 90.08 & 90.23 & 90.15 & 91.69 & 85.41 & 85.83 & 85.62 & 81.22 & 75.87 & 78.46 & 82.04 \\
		\hdashline
        \rowcolor{myrowblue}
		\textbf{Average} & 93.02 & 93.69 & 93.36 & 90.01 & 90.13 & 90.07 & \flatAvg{91.71} & 85.75 & 87.21 & 86.47 & 81.17 & 76.71 & 78.87 & \nestedAvg{82.67} \\
		\midrule
		\addlinespace[1pt]
		\rowcolor{gray!20}
		\multicolumn{15}{c}{\textit{Large Language Models (Category-grouped JSON Format)}} \\
		LLaMA3.1-8B & 92.08 & 93.04 & 92.56 & 89.87 & 89.82 & 89.84 & 91.20 & 85.69 & 86.59 & 86.13 & 79.77 & 76.87 & 78.29 & 82.21 \\
		LLaMA3.2-3B & 92.18 & 92.63 & 92.41 & 87.12 & 87.56 & 87.34 & 89.88 & 83.39 & 83.99 & 83.69 & 79.36 & 76.68 & 77.99 & 80.84 \\
		Qwen2.5-7B & 92.44 & 93.04 & 92.74 & 89.49 & 89.46 & 89.48 & 91.11 & 84.79 & 84.88 & 84.84 & 79.37 & 76.97 & 78.15 & 81.50 \\
		Qwen3-4B & 92.28 & 92.67 & 92.47 & 87.18 & 88.18 & 87.67 & 90.07 & 83.74 & 84.98 & 84.35 & 79.86 & 77.14 & 78.48 & 81.42 \\
		Granite3.3-8B & 92.73 & 93.22 & 92.97 & 88.74 & 89.22 & 88.98 & 90.98 & 85.61 & 86.85 & 86.23 & 80.14 & 76.87 & 78.47 & 82.35 \\
		Granite4-3B & 91.48 & 92.07 & 91.78 & 88.51 & 88.37 & 88.44 & 90.11 & 84.02 & 85.54 & 84.77 & 79.79 & 77.03 & 78.38 & 81.58 \\
		Gemma2-9B & 93.18 & 93.79 & 93.48 & 90.55 & 90.53 & 90.54 & 92.01 & 86.40 & 86.88 & 86.64 & 80.09 & 77.81 & 78.93 & 82.79 \\
		Gemma3-4B & 92.36 & 92.71 & 92.53 & 89.19 & 89.16 & 89.17 & 90.85 & 84.54 & 84.48 & 84.51 & 78.83 & 75.60 & 77.18 & 80.85 \\
		\hdashline
        \rowcolor{myrowblue}
		\textbf{Average} & 92.34 & 92.90 & 92.62 & 88.83 & 89.04 & 88.93 & \flatAvg{90.78} & 84.77 & 85.52 & 85.14 & 79.65 & 76.87 & 78.23 & \nestedAvg{81.69} \\
		\midrule
		\addlinespace[1pt]
		\rowcolor{gray!20}
		\multicolumn{15}{c}{\textit{Large Language Models (Occurrence-based JSON Format)}} \\
		LLaMA3.1-8B & 92.89 & 94.19 & 93.54 & 90.27 & 90.76 & 90.52 & 92.03 & 82.78 & 84.06 & 83.41 & 80.07 & 77.30 & 78.66 & 81.04 \\
		LLaMA3.2-3B & 92.96 & 94.21 & 93.58 & 87.23 & 88.05 & 87.64 & 90.61 & 80.27 & 81.59 & 80.93 & 78.68 & 76.19 & 77.41 & 79.17 \\
		Qwen2.5-7B & 93.10 & 93.94 & 93.52 & 89.77 & 90.32 & 90.04 & 91.78 & 81.46 & 81.92 & 81.69 & 78.60 & 76.58 & 77.58 & 79.64 \\
		Qwen3-4B & 92.26 & 93.56 & 92.91 & 87.51 & 88.45 & 87.98 & 90.45 & 80.85 & 82.31 & 81.58 & 79.32 & 77.63 & 78.47 & 80.03 \\
		Granite3.3-8B & 92.30 & 93.59 & 92.94 & 88.84 & 89.57 & 89.20 & 91.07 & 82.08 & 83.99 & 83.02 & 79.67 & 76.93 & 78.27 & 80.65 \\
		Granite4-3B & 92.29 & 93.24 & 92.76 & 88.31 & 88.94 & 88.63 & 90.70 & 80.84 & 82.94 & 81.88 & 78.55 & 75.99 & 77.25 & 79.57 \\
		Gemma2-9B & 92.99 & 94.12 & 93.55 & 90.08 & 90.50 & 90.29 & 91.92 & 83.49 & 85.11 & 84.29 & 80.30 & 77.96 & 79.12 & 81.71 \\
		Gemma3-4B & 92.74 & 93.61 & 93.17 & 88.60 & 88.37 & 88.49 & 90.83 & 81.43 & 81.72 & 81.58 & 79.25 & 73.06 & 76.03 & 78.81 \\
		\hdashline
        \rowcolor{myrowblue}
		\textbf{Average} & 92.69 & 93.81 & 93.25 & 88.83 & 89.37 & 89.10 & \flatAvg{91.17} & 81.65 & 82.95 & 82.30 & 79.30 & 76.45 & 77.85 & \nestedAvg{80.07} \\
        
		\midrule
		\addlinespace[1pt]
		\rowcolor{gray!20}
		\multicolumn{15}{c}{\textit{Large Language Models (Offset-based JSON Format)}} \\
		LLaMA3.1-8B & 62.91 & 63.53 & 63.22 & 52.99 & 53.63 & 53.31 & 58.27 & 35.53 & 36.17 & 35.85 & 35.16 & 34.09 & 34.62 & 35.24 \\
		LLaMA3.2-3B & 56.01 & 56.25 & 56.13 & 31.06 & 31.34 & 31.20 & 43.67 & 27.03 & 27.59 & 27.31 & 26.17 & 25.43 & 25.79 & 26.55 \\
		Qwen2.5-7B & 60.52 & 61.17 & 60.84 & 50.67 & 51.17 & 50.92 & 55.88 & 32.54 & 32.64 & 32.59 & 32.52 & 31.53 & 32.02 & 32.31 \\
		Qwen3-4B & 60.86 & 61.51 & 61.18 & 38.03 & 38.66 & 38.34 & 49.76 & 35.06 & 35.78 & 35.41 & 37.51 & 36.14 & 36.81 & 36.11 \\
		Granite3.3-8B & 64.49 & 65.09 & 64.79 & 43.78 & 44.38 & 44.08 & 54.44 & 36.48 & 37.33 & 36.90 & 40.06 & 39.03 & 39.54 & 38.22 \\
		Granite4-3B & 60.99 & 61.81 & 61.40 & 38.71 & 39.10 & 38.91 & 50.16 & 33.00 & 33.60 & 33.30 & 37.55 & 36.08 & 36.80 & 35.05 \\
		Gemma2-9B & 65.21 & 65.86 & 65.53 & 61.15 & 61.75 & 61.45 & 63.49 & 40.45 & 41.16 & 40.80 & 41.75 & 40.55 & 41.15 & 40.98 \\
		Gemma3-4B & 59.67 & 60.02 & 59.85 & 51.50 & 51.19 & 51.35 & 55.60 & 31.67 & 31.52 & 31.60 & 33.56 & 30.57 & 32.00 & 31.80 \\
		\hdashline
        \rowcolor{myrowblue}
		\textbf{Average} & 61.33 & 61.91 & 61.62 & 45.99 & 46.40 & 46.20 & \flatAvg{53.91} & 33.97 & 34.47 & 34.22 & 35.53 & 34.18 & 34.84 & \nestedAvg{34.53} \\
		\bottomrule
	\end{tabular}}
	\caption{Complete Precision, Recall, and F1 results of pre-trained models, GPT-3, and open-source LLMs on flat and nested NER datasets.}
	\label{table:full_prf_results}
\end{table*}

\end{document}